\documentclass[10pt,twocolumn,letterpaper]{article}

\usepackage{iccv}
\usepackage{times}
\usepackage{epsfig}
\usepackage{graphicx}
\usepackage{amsmath}
\usepackage{amssymb}
\usepackage{flushend}

\usepackage[pagebackref=true,breaklinks=true,letterpaper=true,colorlinks,bookmarks=false]{hyperref}

\iccvfinalcopy % *** Uncomment this line for the final submission

\def\iccvPaperID{289} % *** Enter the ICCV Paper ID here

\newcommand{\vect}[1]{{\bf {#1}}}

\ificcvfinal\fi
\begin{document}

%%%%%%%%% TITLE
\title{CDVAE: Co-embedding Deep Variational Auto Encoder \\
for Conditional Variational Generation}

\author{Jiajun Lu, Aditya Deshpande, David Forsyth\\
University of Illinois at Urbana Champaign\\
\{jlu23, ardeshp2, daf\}@illinois.edu
}

\maketitle
%\thispagestyle{empty}

%%%%%%%%% ABSTRACT
\begin{abstract}
Problems such as predicting a new shading field ($Y$) for an image ($X$) are ambiguous: many very distinct solutions are good. 
Representing this ambiguity requires building a conditional model
$P(Y|X)$ of the prediction, conditioned on the image.  Such a model is difficult to train, because we do not usually have
training  data containing many different shadings for the same image. As a result, we need different training examples
to share data to produce good models.  This presents a danger we call ``code space collapse'' --- the training procedure
produces a model that has a very good loss score, but which represents the conditional distribution poorly.  
We demonstrate an improved method for building conditional models by exploiting a metric constraint on training data
that prevents code space collapse.  We demonstrate our model on two example tasks using real data: image saturation
adjustment,  image relighting. We describe quantitative metrics to evaluate ambiguous generation results.  
Our results quantitatively and qualitatively outperform different strong baselines.
\end{abstract}

%%%%%%%%% BODY TEXT
\section{Introduction}
\label{sec:intro}

\begin{figure}[ht]
\centerline{  \includegraphics[width=0.5 \textwidth]{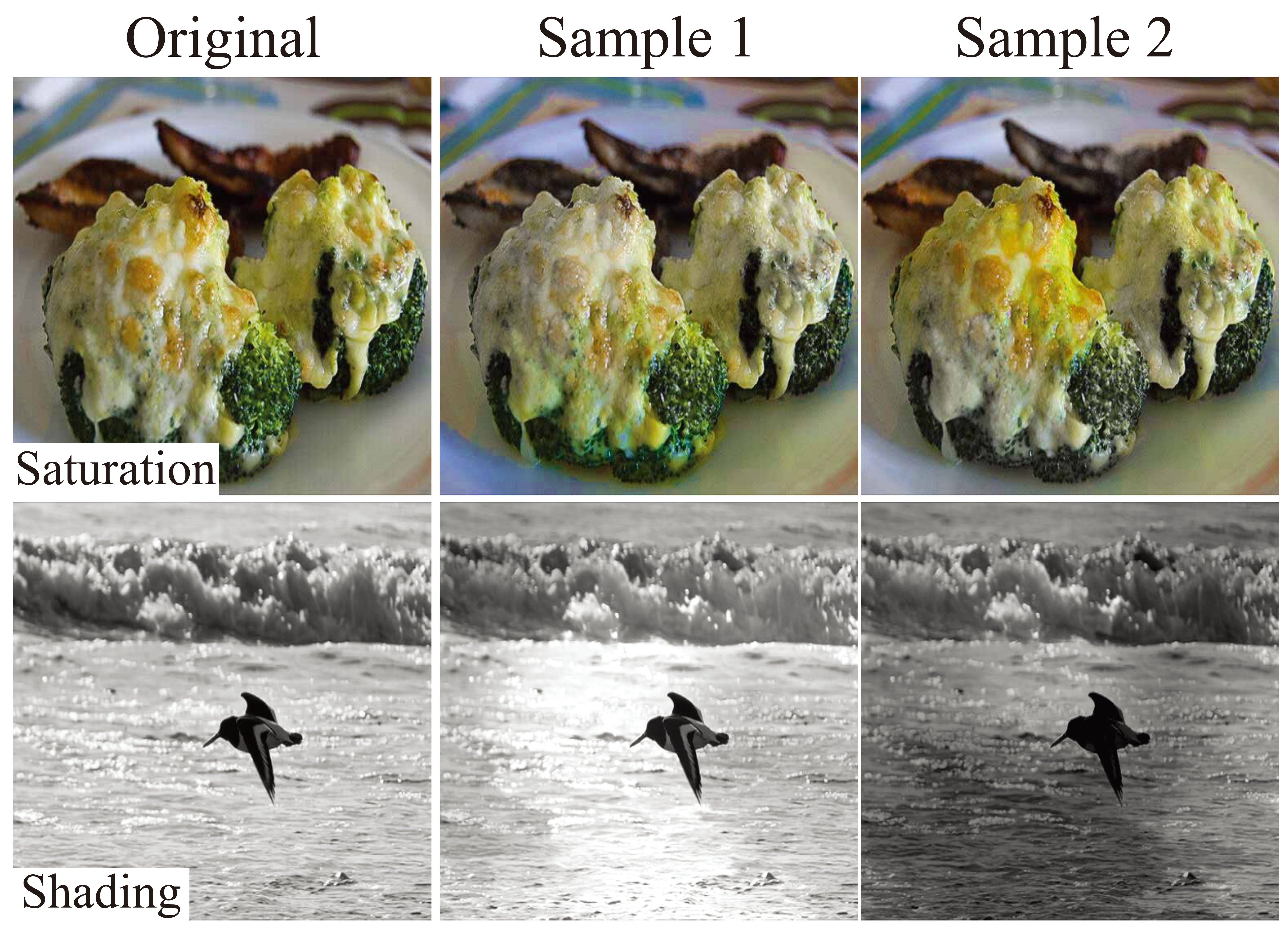}}
  \caption{We describe a method to learn a multimodal conditional distribution $P(Y|X)$ between output spatial field $Y$
    and an input image $X$. We learn the model from ``scattered'' data, where in training one never sees two distinct $Y$ values 
    for one particular $X$ value.  Our method allows us to sample and produce new saturation and shading fields for input images.}
  \label{fig:teaser}
\end{figure}

Many vision problems have ambiguous solutions.  There are many motion
fields consistent with an image~\cite{sohn2015learning,walker2016uncertain,xue2016visual}. 
Similarly, there are many shading fields consistent with the layout of an image 
(Figure \ref{fig:teaser}); many good ways to colorize an 
image~\cite{DivColor, UnsupDivColor, zhang2016colorful}; many possible ways to adjust 
the saturation of an image (Figure \ref{fig:teaser}); many possible long term futures for an image frame~\cite{zhou2016learning}; and
so on. For each of these problems, one must output  a spatial field $Y$ for an input image 
$X$; but $Y$ is not uniquely determined by $X$.   Worse, $Y$ has complex spatial 
structure (for example, saturation at a pixel is typically similar to the saturation at the 
next pixel, except over boundaries). It is natural to build a generative model 
of $Y$, conditioned on $X$, and draw samples from that model. Towards this end, recent 
work has modified a strong generative model for images that uses latent variables, the 
variational auto-encoder (VAE) of~\cite{kingma2013auto}, to produce a conditional VAE 
(CVAE) ~\cite{sohn2015learning,walker2016uncertain,xue2016visual}.  The very high 
dimension and complex spatial covariances of $Y$ are managed by the underlying 
latent variable model in the CVAE.  

However, building a good conditional model still poses some challenges.  In most practical vision
problems, the training dataset we can access is ``scattered'', and the model is multimodal. 
Scattered training data consists of pairs of $(y_i, x_i)$, but we never see multiple 
different values of $y_i$ for one particular $x_i$.  Practical vision models compute 
some intermediate representation $c(x)$ (which we call the ``code''), and predict 
$y$ from that intermediate representation and a random (or latent) variable 
$u$. Write $y=F(u; c(x))$.  The hope is that different choices of the random 
variable will produce different values of $y$, and we can therefore predict the entire 
gamut of outputs (shading, motion field, saturation etc.). This implies that the method 
can represent a multimodal $P(Y|X)$.

\begin{figure}[!t]
\centerline{  \includegraphics[width=0.5 \textwidth]{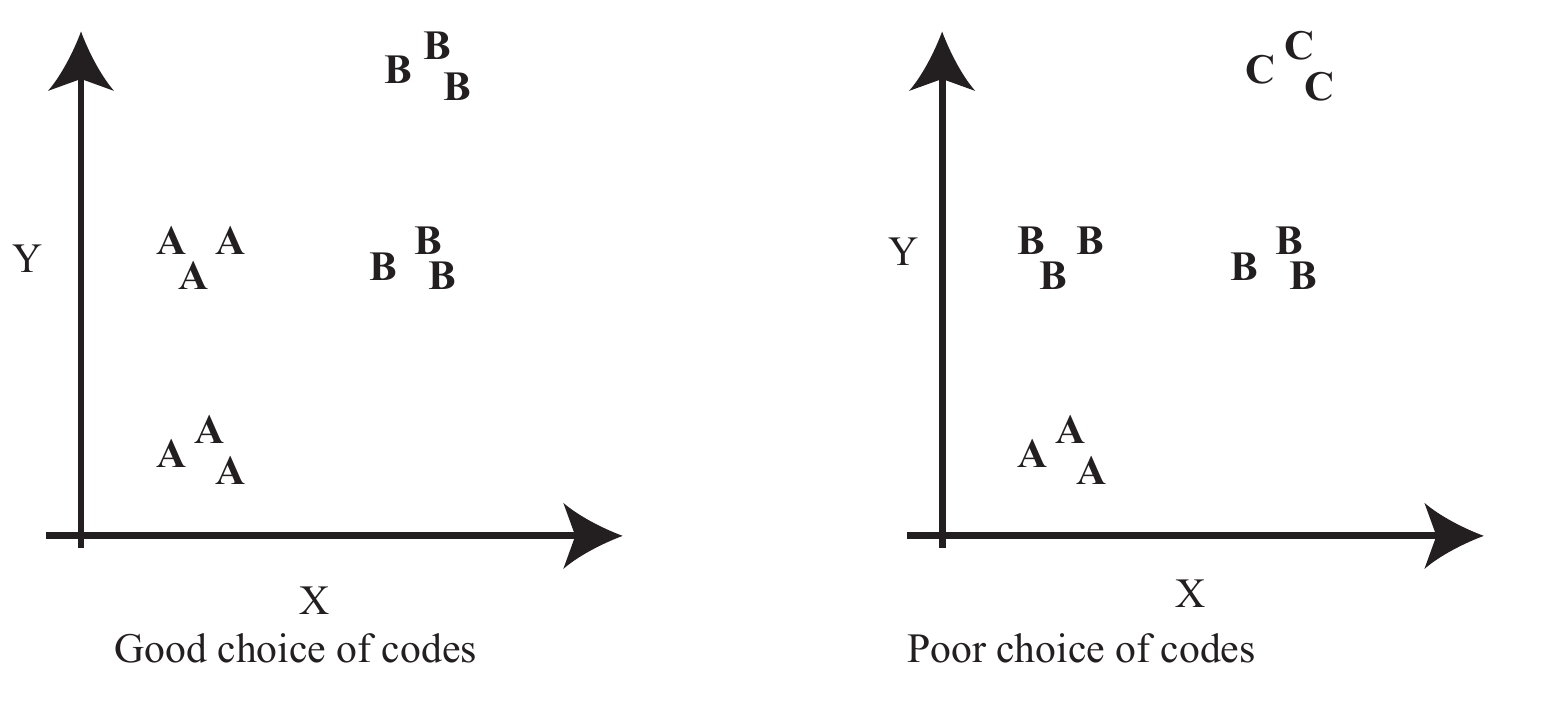}}
  \caption{The horizontal axis is the input $x$, vertical axis is the output $y$ and
  labels A, B and C are the codes $c(x)$. With a good choice of code (left), $F$ is 
  forced to use random variable $u$ to produce different $y$ for similar $x$. With a
  bad choice of code (right), $F$ can choose to ignore $u$ since it has access to 
  different codes for similar $x$. This results in incorrect smoothing and 
  generalization.}
  \label{codes}
\end{figure}

This setting presents a difficulty.  To build a plausible conditional model, we must smooth.
The model smooths by allowing examples with similar {\em codes} to ``share'' $y$ values.  In turn, the choice of code
is crucial, and a poor choice of code can result in a method that appears to work well, but does not.  
Figure~\ref{codes} illustrates this point.   The figure shows two possible choices of code for a particular dataset.  
In the good choice of code, $F$ is forced to use the random variable to produce distinct values of $y$ for similar $x$,
{\em because} similar $x$ result in similar codes.  
%In the poor choice of code, very different $x$'s can have the same code B.  
In the poor choice of code, similar $x$ can have different codes 
(viz. different codes A and B for similar $x$ in the right side of Figure \ref{codes}) and vice versa.  
Then, $F$ can largely ignore the random variable, {\em but} simulates a multimodal process 
by producing very different $y$ for quite similar $x$ using the different $c(x)$.
This means the network we have trained will (a) not be effective at making diverse predictions and (b) may change 
its prediction very significantly for a small change in $x$, in a manner that is uncontrolled by the code.  This is not 
desirable behavior.  

We call the effect ``code collapse'', because the network is
encouraged to produce similar codes for different inputs.  The result is a model with imperfect diversity 
and generalization, but good loss on scattered training data. The absence of variance, in what should be
a diverse pool of output predictions, is a strong diagnostic indicator of code collapse. We exploit 
this indicator and show that our baselines, particularly the CVAE, generate low variance and therefore 
suffer from code collapse (Section ~\ref{sec:quant}, Figure \ref{fig:ev_combine}).

%Such a model may appear to produce some 
%diversity (or variance) in its output, but careful evaluation of diversity confirms the effect occurs 
%(Section~\ref{sec:result}).  

The key problem, resulting in code collapse, is that the current training procedures have no term to force a good choice of 
codes. For example, VAE loss requires the code distribution to look like a standard normal 
distribution. This loss does not force it to preserve the similarity, dissimilar input 
images can be closer and similar inputs further apart in the code space. Recent work shows that better generative models 
are obtained by conditioning on text-embeddings~\cite{Scott16} or pre-trained features from visual 
recognition network~\cite{NguyenYBDC16}. This suggests that using an embedding with some structure
is better than conditioning on raw pixels (with high-capacity networks).  

In our approach, instead of using a fixed embedding as input, we use raw pixels but guide 
the codes (or intermediate representation) with a metric constraint. Our metric term encourages codes $c(x)$ for 
distinct $x$ to be different and codes for similar $x$ to be similar. This prevents code 
collapse. To ensure that $F(u; c(x))$ will vary for similar $c(x)$, we use a Mixture Density 
Network (MDN) \cite{bishop1994mixture}. MDN explicitly models a multimodal conditional 
distribution. We call our model CDVAE (Co-Embedding Deep Variational Auto Encoder).

We apply CDVAE to two novel (from the point of view of automated editing) problems: (a) Photo Relighting, 
(b) Image Resaturation (Section \ref{sec:apps}). In relighting (or reshading), we decompose the image into shading and albedo, then produce a new shading 
field consistent with the albedo field. In resaturation, we produce a new saturation field and 
apply it to the image.  In each case, the resulting image should look ``natural'' -- like 
a real image, but differently illuminated (reshading) or with differently color saturated objects 
(resaturation).  In all cases, our model outperforms strong baselines 
(including the CVAE). \\

%Our approach conveniently extends to multiple layers of latent variables, and we use the Deep VAE
%of~\cite{rezende2014stochastic} for our problems. 

\begin{figure*}[ht]
\centerline{  \includegraphics[width=1.0 \textwidth]{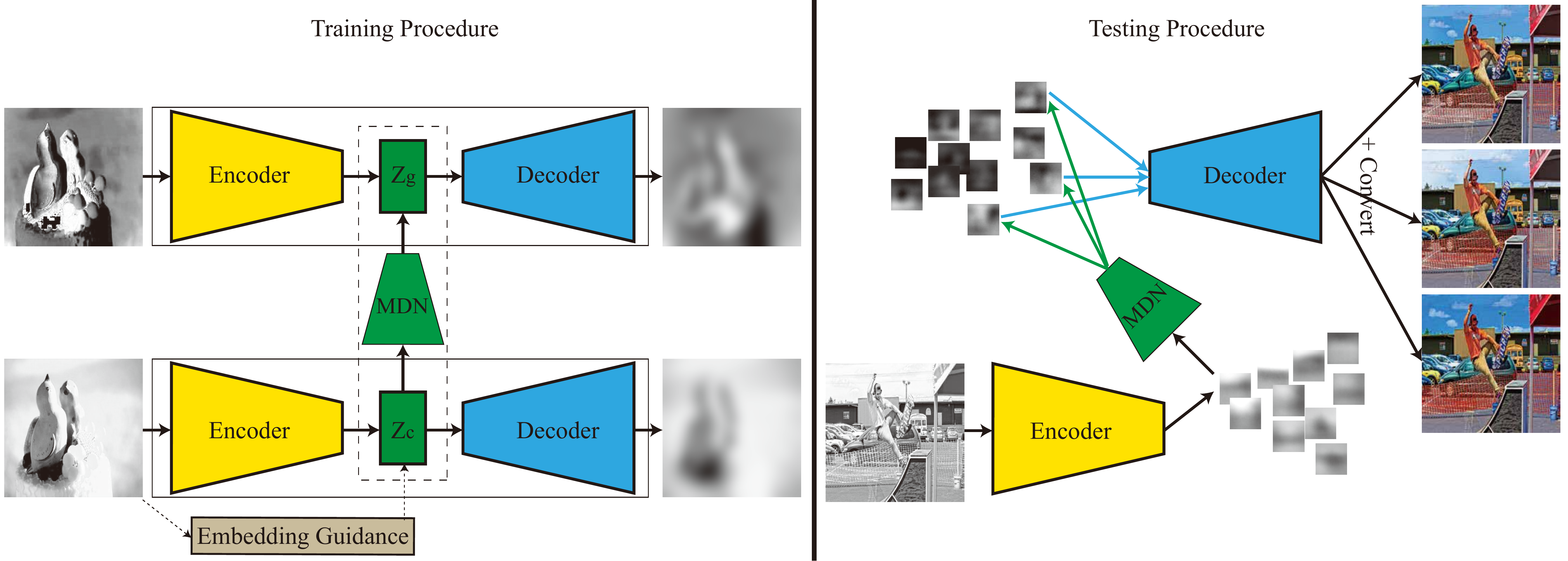}}
  \caption{{\bf Left} Our training architecture for CDVAE; and {\bf Right} Test-time architecture
   of CDVAE. We use two deep variational autoencoders (DVAE), one for the conditioning image $x_c$ and 
   another for generated image $x_g$. Each DVAE has two layers of latent gaussian variables and we 
   use the ladder VAE architecture of \cite{sonderby2016ladder}. Embedding guidance introduces
   metric constraints on the code space $z_c$ to prevent code collapse. And, MDN models the multimodal
   distribution between $z_g$ and $z_c$. During test, we sample multiple $z_g$ from MDN for a given 
   input. We decode these different $z_g$ to obtain multiple predictions.}
  \label{fig:architecture}
\end{figure*}

%\vspace{2mm}
{\bf Contributions:}
\begin{itemize}
\item We describe a novel method to build conditional models for extremely demanding 
datasets (Section \ref{sec:method}, Figure \ref{fig:architecture}) and apply it to
photo-editing tasks (Section \ref{sec:apps}).  
\item We show how to regularize our model so that the latent representation is not distorted, and this helps us 
improve results (Section \ref{sec:embed} and Figure \ref{fig:ev_combine}). 
\item Our method is compared to a variety of strong baselines, and produces predictions that (a) have high variance and
  (b) have high accuracy (Section \ref{sec:baseline} and Section \ref{sec:quant}). Our method clearly outperforms existing models.
\item Training previous conditional models is hard, these models tend to either go to code collapse 
or random prediction. Our methods can avoid code collapse and create multiple distinct plausible results 
(Section \ref{sec:qual} and Figure \ref{fig:cmp}). 
\end{itemize}

\section{Related Work}

Generating a spatial field with complex spatial structure from an image is an established problem. Important application
examples, where the prediction is naturally ambiguous, include
colorization~\cite{deshpande2015learning,larsson2016learning, zhang2016colorful}, style transfer~\cite{gatys2015neural},  
temporal transformations prediction~\cite{zhou2016learning}, predicting
motion fields~\cite{sohn2015learning,walker2016uncertain,xue2016visual}, and predicting future
frames~\cite{vondrick2015anticipating}.  
This is usually treated as a regression problem; current state of the art methods use deep networks to learn features. 
However, predicting the expected value of a conditional distribution, through regression, works poorly, because the expected value of a
multimodal distribution may have low probability.   While one might temper the distribution
(eg~\cite{zhang2016colorful}), the ideal is to obtain multiple different solutions.  

One strategy is to draw samples from a generative model.  Generative models of images present problems of dimension;
these can be addressed with latent variable models, leading to the variational autoencoder (VAE) of
\cite{kingma2013auto}.   As the topic attracts jargon, we review the standard VAE briefly here. This approach learns an
encoder $E$ that maps data $x$ into a continuous latent variable $z=E(x)$ (the codes), and a decoder $D$ that maps $z$ to
an image $\hat{x}=D(z)$.  Learning ensures that (a) $x$ and $D(E(x))$ are close; (b) if $\xi$ is close to $z$, then
$D(\xi)$ is close to $D(z)$; and (c) $z$ is distributed like a standard normal random variable.   Images can then be
generated by sampling a standard normal distribution to get $\xi$, and forming $D(\xi)$.  This is a 
latent variable model, modelling $P(x)$ as $P(x) = \int_z P(x|z)P(z)dz$. $P(x|z)$ is represented by the decoder,
we use an auxiliary distribution $Q = P(z|x)$ for $P(z)$, where $P(z|x)$ is now the encoder.  Learning is by maximizing a 
lower-bound on log-likelihood (Equation \ref{eq:vae}) 
\begin{equation}
\mbox{VAE}(\theta) = \sum_{data}\mathbb{E}_Q [\log P(x|z)] - \mathbb{D}(Q || P(z))
\label{eq:vae}
\end{equation}
where $\theta$ are the parameters of encoder and decoder networks of the VAE.

Current generative models are reliable in simple domains (handwritten digits~\cite{kingma2013auto, salimans2015markov};
faces~\cite{kingma2013auto, kulkarni2015deep, rezende2014stochastic}; and CIFAR images~\cite{gregor2015draw}) but less
so for general images.  Improvements are available using multiple layers of latent variables (a
DVAE)~\cite{rezende2014stochastic}. Models can be trained with adversarial loss~\cite{radford2015unsupervised}. However, these deep
models are still hard to train. The ladder VAE imposes both top-down and bottom up  variational distributions for more
efficient training~\cite{sonderby2016ladder}. 

The generative model needs to be conditioned on an image, and needs to be multimodal. Tang et al. give a multimodal conditional
model~\cite{tang2013learning}, but the conditioning variables are binary. A conditional variational autoencoder
(CVAE)~\cite{sohn2015learning} conditions the decoder model on a continuous representation produced by a network applied to $x$.
This approach has been demonstrated on motion prediction problems~\cite{walker2016uncertain,zhou2016learning}. 

We use the mixture density network (MDN) in our models to capture the underlying multimodal distribution~\cite{bishop1994mixture}. 
MDN predicts the parameters of a mixture of gaussians from real-valued input features. MDNs have been successfully applied to
articulatory-acoustic inversion mapping~\cite{richmond2007trajectory, uria2012deep} and speech synthesis~\cite{zen2014deep}.

\section{Method}
\label{sec:method}

Our CDVAE consists of two deep variational auto encoders (DVAE)~\cite{rezende2014stochastic} 
and a mixture density network (MDN)~\cite{bishop1994mixture}. An overview of the CDVAE model 
is shown in Figure~\ref{fig:architecture}. 

Our architecture is, we use two DVAEs to embed the conditioning image $x_c$ 
(or $X$ from Section \ref{sec:intro}) and the generated image $x_g$ 
(or $Y$ from Section \ref{sec:intro}) into two low-dimensional latent
variables (or code spaces) $z_c$ and $z_g$. The generated image  
corresponds to output spatial field viz.\ saturation or shading etc.\ and 
the conditioning image is the input image viz.\ intensity or albedo etc.\ 
Next, we regularize the latent variables $z_c$ with embedding guidance 
(or metric constraints) such that the similarity in input space is maintained 
(Section \ref{sec:embed}). Since our problem is ambiguous, the conditional 
distribution between $z_g$ and $z_c$ is multimodal. MDN allows us to fit a multimodal gaussian 
mixture model (GMM) between the conditioning code $z_c$ and the generated 
code $z_g$ (Section \ref{sec:mdn}). At test time, we sample from this multimodal 
GMM ($z_g \sim MDN$) and use the decoder of the generated 
image DVAE ($Decoder(z_g)$) to predict different shading, saturation  for 
a single input image.

In Figure \ref{fig:architecture}, we simplify and show a single layer of 
latent variables. In practice, our DVAE utilizes multiple layers of gaussian 
latent variables. This hierarchical variational auto encoder architecture 
captures complex structure in the data, and provides good practical performance. 

We jointly train the two DVAEs and the MDN, allowing them to adapt and mutually benefit each 
other. Joint training also enables CDVAE to model a joint probability distribution
of $x_c$ and $x_g$, instead of a conditional probability distribution (viz.\ like a 
CVAE). The joint probability model allows for more smoothing between data points. 
In CDVAE, we optimize the joint probability model $P(x_c, x_g)$ during training 
(Refer Equation \ref{eq:joint}). At test time, we remove the decoder for the 
conditioning layer and the encoder for the generated layer. This converts the 
joint model of CDVAE into a conditional model $P(x_g|x_c)$. We can then sample this 
conditional model to generate diverse outputs. Similar to VAE probability model, 
we write the joint probability by marginalizing over the joint distribution $P(z_g, z_c)$

\begin{equation}
\medmuskip=1mu
\thinmuskip=1mu
\thickmuskip=1mu
P(x_c, x_g)  = \int_{z_c} \int_{z_g} P(x_c|z_c) P(x_g|z_g)  P(z_g, z_c) dz_g dz_c
\label{eq:joint}
\end{equation}

In Section \ref{sec:mdn}, we derive the loss terms corresponding to this joint 
probability model of CDVAE.

\subsection{CDVAE Loss}
\label{sec:mdn}

In CDVAE, we use two multi-layer variational auto encoders (DVAE), one for $x_c$ and one 
for $x_g$. Additionally, we have an MDN that models the relationship between the embeddings
$z_c$ and $z_g$. The loss function $\mathcal{L}_{CDVAE}$ corresponding to the CDVAE joint model $P(x_{c}, x_{g})$ 
is a combination of the two DVAE models and the conditional probability model of MDN. 
In Equations~\ref{eq:vae} and ~\ref{eq:joint}, 
assume it is possible to encode $x_c$ without seeing $x_g$, then we can use the auxiliary 
sampling distribution $Q=P(z_c|x_c)P(z_g|x_g)$ for our CDVAE. If $P(z_g, z_c)=P(z_g) P(z_c)$ 
in Equation~\ref{eq:joint}, we can separate out the product terms of joint probability 
model. Taking negative log-likelihood, we obtain separate additive loss terms for each 
DVAE. Write $\mbox{DVAE}(\theta)$ for the loss function of a DVAE with weights
(or parameters) $\theta$, which has the standard form of a VAE loss (Equation \ref{eq:vae}). 
Write the loss of the DVAE for the generated image $x_g$ as $\mbox{DVAE}(\theta_g)$ 
(similarly for $x_c$). Write loss for the MDN as $\mathcal{L}_{mdn}$. We can then derive the 
loss function for CDVAE (Equation \ref{eq:cdvae_loss}), 
\begin{equation}
\mathcal{L}_{CDVAE} =  \mbox{DVAE}(\theta_c) + \mbox{DVAE}(\theta_g) + \mathcal{L}_{mdn}(\theta_c, \theta_g)
\label{eq:cdvae_loss}
\end{equation}

Our MDN estimates the conditional probability model $P(z_{g} | z_{c})$. For each input code 
$z_c$, our MDN estimates the parameters of a $K$ component gaussian mixture distribution with 
mixture coefficients $\pi_i$, means $\mu_i$ and fixed diagonal covariance $\sigma_i^2$. The
loss $\mathcal{L}_{mdn}$ is obtained by taking the negative log-likelihood of this 
conditional distribution, 

\begin{equation}
\medmuskip=1mu
\thinmuskip=1mu
\thickmuskip=1mu
%\resizebox{1.0\hsize}{!}{$  \mathcal{L}_{mdn} = -E_{Q}\left[\log \sum_{k=1}^K \pi_k(z_c) \mathcal{N}(z_g|\mu_k(z_c), \sigma_k^2(z_c))\right]  $}
 \mathcal{L}_{mdn} = -E_{Q}\left[\log \sum_{k=1}^K \pi_k(z_c) \mathcal{N}(z_g|\mu_k(z_c), \sigma_k^2(z_c))\right]  
\end{equation}

We use an inference method~\cite{sonderby2016ladder} that co-relates the latent variables of 
the encoder and the decoder. This speeds up the training. Refer to the supplementary materials 
for the detailed derivations.

%to obtain
%\begin{equation}
%\mathcal{L}_{CDVAE, mdn} =  \mbox{DVAE}(\theta_c) + \mbox{DVAE}(\theta_g) + \mathcal{L}_{mdn}(\theta_c, \theta_g).
%\end{equation}
%Since the joint model linking $z_c$ and $z_g$ is a mixture of normal distributions,
%$P(z_g|z_c)$ is a mixture of normal distributions, too. This means the resulting conditional model is also easy to sample. 
% During the generative process, we draw a sample by sampling the mixture model $\sum_{k=1}^K \pi_k(z_c) \mathcal{N}(z_g|\mu_k(z_c), \sigma_k^2(z_c))$. 
%The multiple modes property enables MDN "tie" to search in different local manifolds and creates more various results.

\subsection{Embedding Guidance: Preventing code collapse}
\label{sec:embed}

\begin{figure}[!t]
\centerline{  \includegraphics[width=0.5 \textwidth]{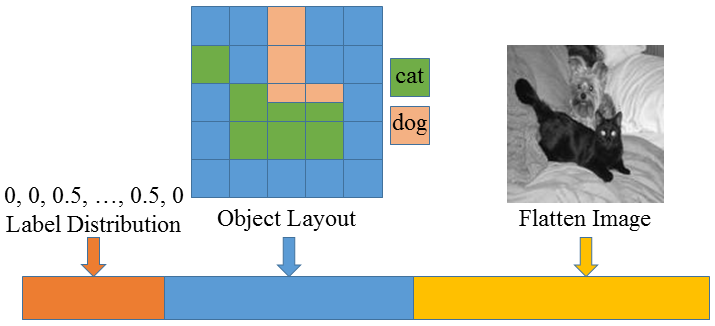}}
%\vspace{2 mm}
  \caption{We use Niyogi~\cite{niyogi2004locality} to build an embedding which 
   preserves metric relations between data points. Our input feature vector to~\cite{niyogi2004locality} 
   is constructed in three parts. First part is the semantic label distribution, which describes the 
   object label percentages in the image (viz.\ 0.5 cat and 0.5 dog etc.). Second part is the object 
   layout, which includes information of spatial layout percentages. The last part is resized 
   image pixels, which provides low level image information.}
  \label{fig:fea}
\end{figure}

Vanilla DVAEs have difficulty in learning a code space which encodes the complex spatial 
structure of the output. The code space learned by a DVAE appears to be underdetermined, 
especially for large and complex datasets. This is a common failure mode of VAEs. 
For our conditional models, it is desirable that codes for ``similar'' inputs are nearby, 
and codes for ``very different'' inputs are further apart. This discourages the method 
from grouping together very different inputs. It also prevents similar images from
having different codes, and therefore avoids incorrect smoothing of the model (See 
Figure \ref{codes}). We guide the codes (at multiple layers) to be similar to a pre-computed 
embedding. Our pre-computed embedding is such that it preserves the similarity observed in 
input domain. Refer to Figure \ref{fig:fea} and supplementary material for the details 
of our pre-computed embedding. Write $p$ for the pre-computed embedding and $z_c$ for the gaussian latent 
variables of the network. We use the L$_{2}$-norm between $p$ and $z_c$ as a loss term

\begin{equation}
\mathcal{L}_{embed} = \| z_c-p \|_{2}^2
\end{equation}

%Our embedding guidance discourages major distortions of the code space.  Our approach can 
%be seen as a regularization similar to batch normalization~\cite{ioffe2015batch}, which 
%discourages (but does not prevent) large values between layers. We can use the embedding 
%guidance to customize the code space to the applications. 

Write $\mathcal{L}$ for the final loss function with the additional regularization in the 
form of embedding guidance

\begin{equation}
\mathcal{L} = \mathcal{L}_{CDVAE}+ \lambda \mathcal{L}_{embed}
\label{eq:loss}
\end{equation}

We use a large value of $\lambda$ when training starts and gradually reduce it 
during the training process.

\subsection{Post Processing}
\label{sec:postprocess}
Current deep generative models only handle small images, for our case $32 \times 32$, and 
they generate results without high spatial frequencies. We post process generated images 
for viewing at high resolution (not used in any quantitative evaluation). Our post processing 
upsamples results to a resolution of $512 \times 512$, with more details.  We aggressively 
upsample the generated fields with the approach in ~\cite{lu2015sparse}, which preserves edges during 
upsampling. In particular, the method represents a high resolution field by a weight and
an orientation value at each  sample point; these parameters control a basis function placed at each sample, and the
field at any test point is a sum  of the values of the basis functions at that point. 
The sum of basis functions includes only those basis functions in the same image segment. 
Write $\mbox{Interp}(\vect{w}, \theta; {\cal S}(I))$ for the
high resolution field produced by interpolating a weight vector $\vect{w}$ and a vector of orientations $\theta$,
assuming a segment mask ${\cal S}(I)$ obtained from the high resolution grey level image.   Write $Y^{(d)}$ for a low
resolution field produced by the decoder, and $\downarrow G_\sigma *$ for a process that smooths and
downsamples.  We solve
\[
\downarrow G_\sigma *(\mbox{Interp}(\vect{w}, \theta; {\cal S}(I))) = Y^{(d)}
\]
for $\vect{w}$, $\theta$, regularizing with the magnitude of $\vect{w}$.  This produces a high resolution field that (a) is compatible with
the high resolution edges of the grey level image (unlike the learned upsampling in common use) and (b) produces the decoder sample when smoothed.

For relighting and saturation adjustment tasks, we polish images for display by using detail maps to recover fine scale
details. The detail map is calculated by taking the conditioning image $I_c$ and subtracting the output 
produced by the conditioning image decoder with the code $z_{c}$.  This captures the high frequency details 
lost during the neural network process. We get our result $\hat{I_g}$ as
\begin{equation}
\hat{I_g} = \mbox{CDVAE}(z_g) +(I_{c} - \mbox{CDVAE}(z_{c}))
\end{equation}

\section{Applications}
\label{sec:apps}

We apply our methods to two different ambiguous tasks, each of which admits both quantitative 
and qualitative evaluations.

\begin{figure*}[ht]
\centerline{  \includegraphics[width=0.8 \textwidth]{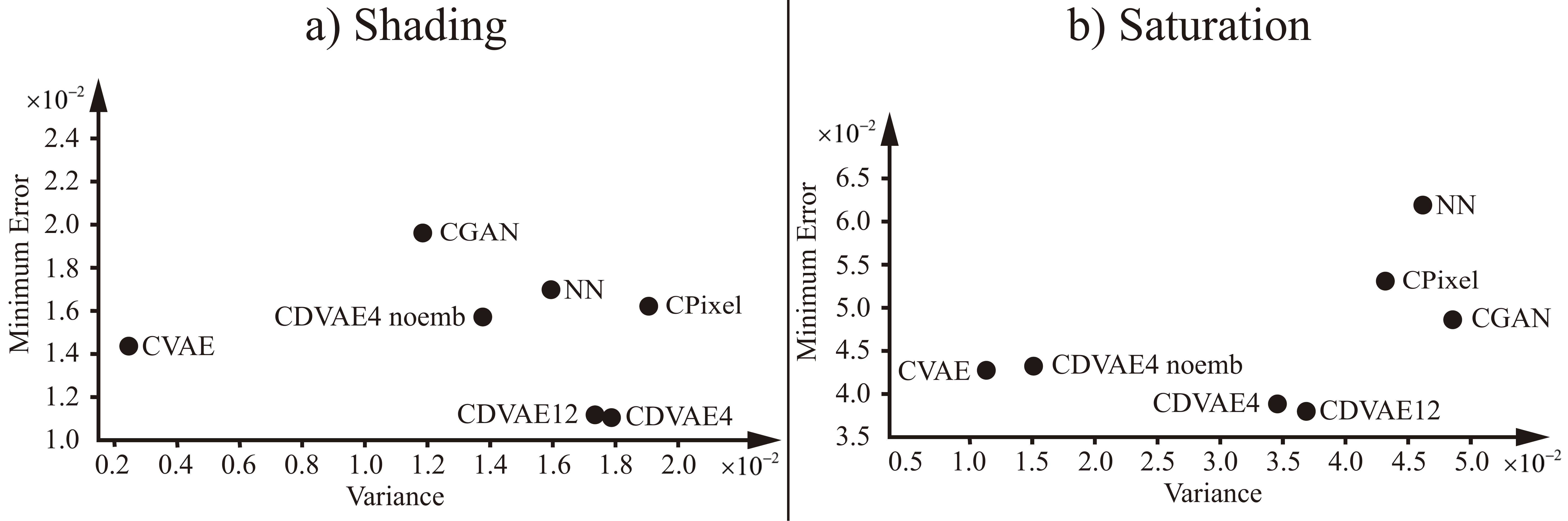}}
%\vspace{2 mm}
  \caption{Comparison to baselines (Section \ref{sec:baseline}) on two tasks: a) photo relighting, b) image resaturation. Vertical axis is 
  error of closest sample to the ground truth, and horizontal axis is variance of predicted samples (bottom right is 
  better).  Both are calculated by sampling 100 outputs from all conditional models. In all tasks, CVAE has low variance for generated results, 
  suggesting the method is poor at producing diverse samples. The nearest neighbor has higher variance but cannot predict samples close to ground-truth
  (higher minimum error). For our CDVAE, the performance increases with 12 MDN gaussian kernals as opposed to 4 and 
  embedding guidance is useful (CDVAE noemb's performance drops). Tables with detailed numbers are in the supplementary materials. CGANs and
  Conditional PixelCNN (CPixel) have higher minimum error, indicating they produce less natural output spatial fields.
}
  \label{fig:ev_combine}
\end{figure*}

{\bf Photo Relighting (or Reshading):} In this application, we predict new shading fields for images which are consistent with a relit
version of the scene.  We decompose the image into albedo (the conditioning image $x_c$), and shading (the generated image $x_g$). 
In real images, shading is quite strongly conditioned by albedo because the albedo image contains semantic information such 
as the scene categories, object layout and potential light sources. A scene can be lighted in multiple ways, so relighting 
is a multimodal problem. We use the MS-COCO dataset for relighting (Section \ref{sec:data}).

% Our CDVAE model has semantic sensing of the scene and can figure out the objects layout, so it is an ideal way to perform objects insertion. In this framework, our
% object insertion operation is performed on the albedo images, and we use our CDVAE to generate a novel consistent shading maps to relight the whole image to 
% make a consistent object insertion results. 

{\bf Image Resaturation:} Here, we predict new saturation fields for color images, i.e.\ we 
modify color saturation for some or all objects in the image.  We transform the input RGB color image into 
HSV color space, and use the value channel as our conditioning image and the saturation channel as our 
generated image.  We use CDVAE to generate new saturation fields consistent with the input value image. 
Combining the new saturation fields with H and V channels leads to new versions of an image, 
with edited saturation. See the grilled cheese on the broccoli in Figure~\ref{fig:teaser} which demonstrates
that we obtain natural edits. Again, we use the MS-COCO dataset for this application (Section \ref{sec:data}).

\section{Results}
\label{sec:result}
\begin{figure*}[ht]
\centerline{  \includegraphics[width=1.0 \textwidth]{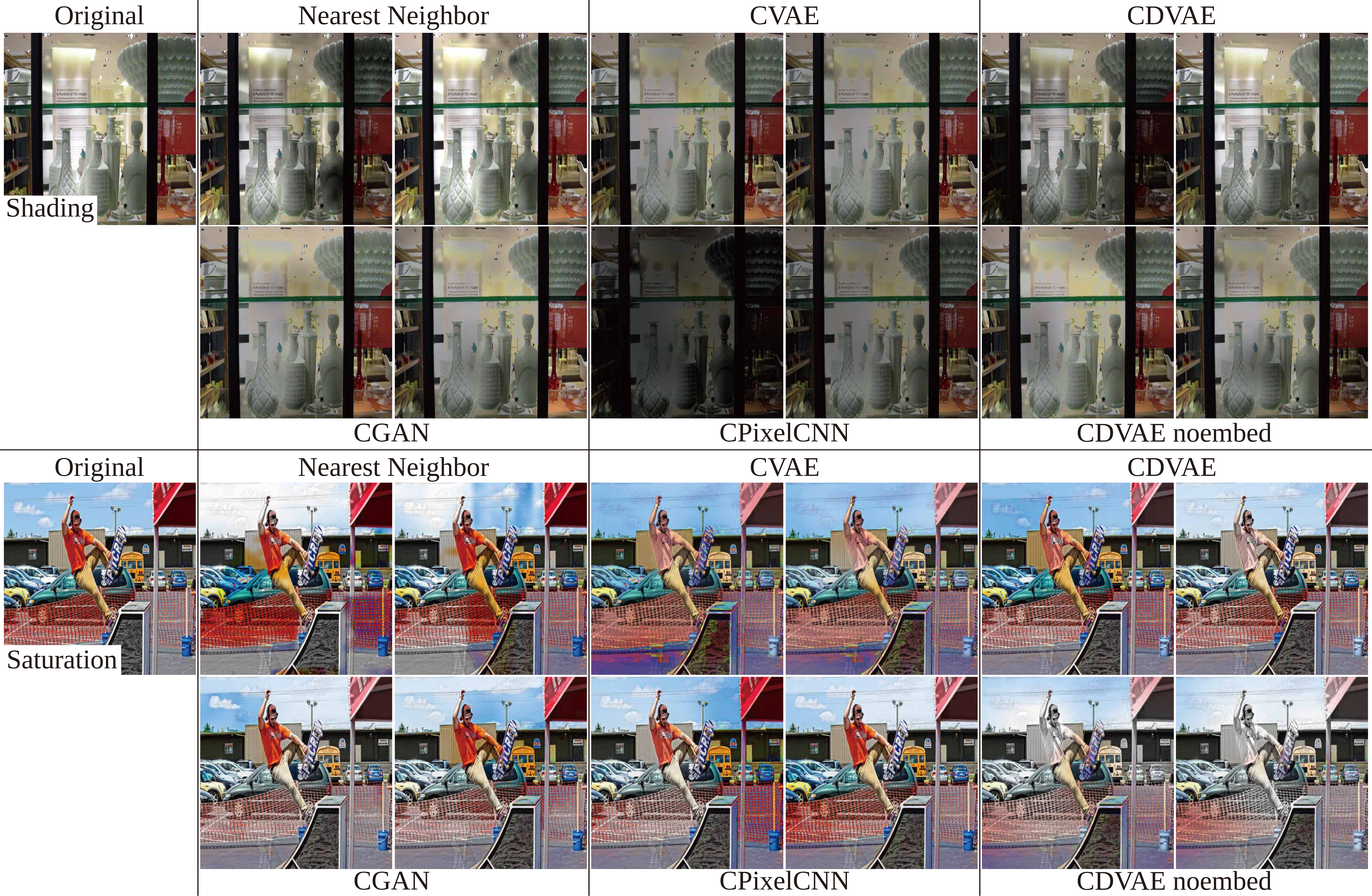}}
%\vspace{2 mm}
  \caption{Qualitative comparisons for Photo Relighting (top) and Image Resaturation (bottom). The first column is the input image. 
%  For tasks of relighting and saturation adjustment, our detail map is applied. If the model keeps outputing a same result with any input, the generated image will be the original image. 
  Nearest neighbor creates inconsistent visual artifacts, since it is a non-parametric method with little awareness
  of the content and spatial structure. CVAE generates low diversity. Notice the diversity in outputs of PixelCNN
  and CGANs is also limited. In contrast, our CDVAE generates two plausible different relighted scenes and it 
  generates two reasonable resaturation outputs (high and low saturation) different from original input . Note that, 
  without the embedding constraint (CDVAE noembed), we observe code collapse and same predictions.}
  \label{fig:cmp}
\end{figure*}

To evaluate the effectiveness of our method, we compare with recent strong methods (Section \ref{sec:baseline}). 
We also evaluate different variants of our method. We perform quantitative and qualitative comparison on 
applications of photo relighting, image resaturation. Quantitative results (Section \ref{sec:quant}, Figure \ref{fig:ev_combine}) are 
computed on the network output, without any post processing.  Images shown for qualitative evaluation of 
resaturation and reshading (Section \ref{sec:qual}, Figures \ref{fig:teaser}, \ref{fig:cmp}, \ref{fig:shading} and \ref{fig:saturation}) 
are post processed using the steps described in Section~\ref{sec:postprocess}. We downsample all the 
input images to $32 \times 32$ dimensions, and all neural network operations before post processing are 
performed on this image size. After our CDVAE model generates samples, our post processing upsamples 
it to a resolution of $512 \times 512$. 

\subsection{Datasets}
\label{sec:data}

\textbf{MS-COCO: } We use MS-COCO dataset for our both the tasks, photo relighting and image resaturation. It is a wild 
dataset (unlike the structured face data commonly used by generative models), and has complex spatial structure. 
Typically, such data is challenging for VAE-based methods. We use train2014 (80K images) for model training, and 
sample 6400 images from val2014 (40K images) for testing. For photo relighting, intrinsic image decomposition method from ~\cite{Bell:2014} 
is used to obtain albedo and shading images. For image resaturation, we transform the image from RGB space to HSV space. 

\subsection{Quantitative Metrics and Evaluation}
\label{sec:quant}

%We perform qualitative and quantitative comparisons on our results of all three tasks. 
%For quantitative evaluation, we report negative loglikelihood, best error to ground truth and variance of sample points. 
%\textbf{Negative log likelihood: } Our implementation of CVAE does not have explicit negative log likelihood, and some previous works use Parzen window method
%to estimate it. However, such estimation is highly unaccurate because of the high dimension, so we do not report our negative loglikelihood for CVAE, and only use the 
%result from motion prediction paper~\cite{walker2016uncertain}. Numbers are reported in Table~\ref{tb:neglog}. For motion prediction, our method and variations outperform 
%CVAE method. Generally, as our model goes deeper and use stronger condtional model, the negative log likelihood becomes smaller. 

\textbf{Error-of-Best to ground truth:} We follow the motion prediction work~\cite{walker2016uncertain} to use 
error-of-best to ground truth as an evaluation metric. We draw 100 samples for each conditional model and calculate 
the minimum per pixel error to ground truth fields. A better model will produce smaller values, because it 
should produce some samples closer to the ground truth.

\textbf{Variance:} 
A key goal of our paper is to generate diverse solutions. However, no current evaluation regime can tell 
whether a pool of diverse solutions is right. We opt for the strictly weaker proxy of measuring variance
in the pool, on the reasonable assumption that diverse predictions for our problems must have high variance.
Thus, procedures that produce low variance predictions are clearly unacceptable. Clearly, it is not enough
just to produce variance -- we want the pool to contain appealing diverse predictions. To assess this, 
we rely on qualitative evaluations (Figures \ref{fig:cmp}, \ref{fig:saturation}, \ref{fig:shading}).
The supplementary materials contain many additional qualitative results.

To compute variance, we obtain the values for 16 ($4 \times 4$) equally spaced grid 
(since distant pixels are de-correlated to some extent) of pixels in our $32 \times 32$ 
prediction. We collect these values across 100 samples, and compute the variance at 
each grid point across samples. We report this average variance vs. minimum error 
(See Figure~\ref{fig:ev_combine}). In particular, a method with more diverse output 
predictions should result in higher variances and one of them should also be close
to the ground-truth (therefore, low minimum error). Specifically, we need to be in
the bottom-right part of Figure \ref{fig:ev_combine}, which our CDVAE achieves.
 
Therefore, our CDVAE model creates results with desirable properties: lower error-of-best 
to ground truth and large variance. Our CDVAE model produces better results with more 
gaussian kernels (CDVAE12 vs. CDVAE4) and performance drops (higher minimum error and 
low variance) when no embedding guidance is used (CDVAE4 noemb). 

\subsection{Baseline Methods}
\label{sec:baseline}

\textbf{Nearest neighbor (NN): } We perform top$-k$ nearest neighbor (NN) search in 
$x_c$ space, and return the $k$ corresponding $x_g$ as multiple outputs. Gaussian 
smoothing is applied to returned $x_g$ to remove inconsistent high frequency 
signal. Since our training data does not have explicit one-to-many correspondences, 
NN is a natural strong baseline. It is a non-parametric method to model multimodal
distribution by borrowing output spatial fields (we also smooth these) from nearby 
input images.

\textbf{Conditional variational autoencoder: } We implement a CVAE similar to 
\cite{walker2016uncertain}. We cannot use \cite{walker2016uncertain} since
their architecture is specific to prediction of coarse motion trajectories. 
Our decoder is modelled on the DCGAN architecture of Radford et al.~\cite{radford2015unsupervised} with
5 deconvolution layers, and we use codes of dimension $64$ (to be consistent
with CDVAE). Our image tower and encoder tower use 5 convolutional layers (mirror
image of the decoder). We use the same strategy as \cite{walker2016uncertain}, 
i.e.\ we spatially replicate code from encoder and multiply it to the output of image
tower. The decoder takes as input the result of this. We train our CVAE with the standard 
pixel-wise L2 loss on output and KL-divergence loss on the code space. At test time, 
codes are randomly sampled from the normal distribution. 

\textbf{Conditional GAN:} CGAN~\cite{mirza2014conditional} is another conditional 
image generation model. It uses a regularized code (drawn from a uniform distribution) 
along with a fixed embedding of the conditioning image as input. We observe that
CGAN achieves higher minimum error (or error-of-best) and lower variance as 
compared to CDVAE (See Figure \ref{fig:ev_combine}). Therefore, we generate better 
(lower minimum error) and more diverse (higher variance) predictions that CGAN.
These metrics are explained in detail in Section \ref{sec:quant}.

%CGAN is first used to generate images from labels, which means the training data has one label to multiple image correspondeces. Then it has been applied to one to
%one correspondence dataset. It usually generate sharper images than CVAE, however it suffers from problems of mode collapsion, training difficulties and so on. 

\textbf{Conditional PixelCNN (CPixel):} Conditional PixelCNN~\cite{oord2016conditional} uses masked
and gated convolutions (sigmoid and tanh activations layers multiplied). The 
receptive field of masked convolutions mimics the causal dependency and gated convolutions
approximate the behavior of LSTM gates in recurrent architectures. Therefore, PixelCNN (CPixel) 
feasibly approximates the compute intensive recurrent architectures~\cite{gregor2015draw} for 
image generation. However, their receptive field grows linearly and handling long-scale effects 
is difficult. Our results are qualitatively better than PixelCNN, we believe our DVAE with 
fully-connected layers is better at capturing the global structure given the coarse resolution 
used. Note, our CDVAE has lower minimum error than PixelCNN (Figure \ref{fig:ev_combine}).

%generates decent images in lots of tasks. It's a strong baseline, and the biggest problem
%of this approach is heavy computation. The computation takes hundreds of times compared to most generative methods in both training and testing process.

\subsection{Qualitative Evaluation}
\label{sec:qual} 

In addition to outperforming baselines on quantitative results, our method generates better
qualitative results. Samples from our jointly trained conditional model smooth information
across ``similar'' images, allowing us to produce aligned, semantically sensible and 
reasonable diverse predictions.  Our qualitative comparisons with other methods for the two tasks 
is shown in Figure ~\ref{fig:cmp}.  In both examples, we generate plausible and diverse relighted 
scenes and resaturated image. 

\begin{figure}[!t]
\centerline{  \includegraphics[width=0.5 \textwidth]{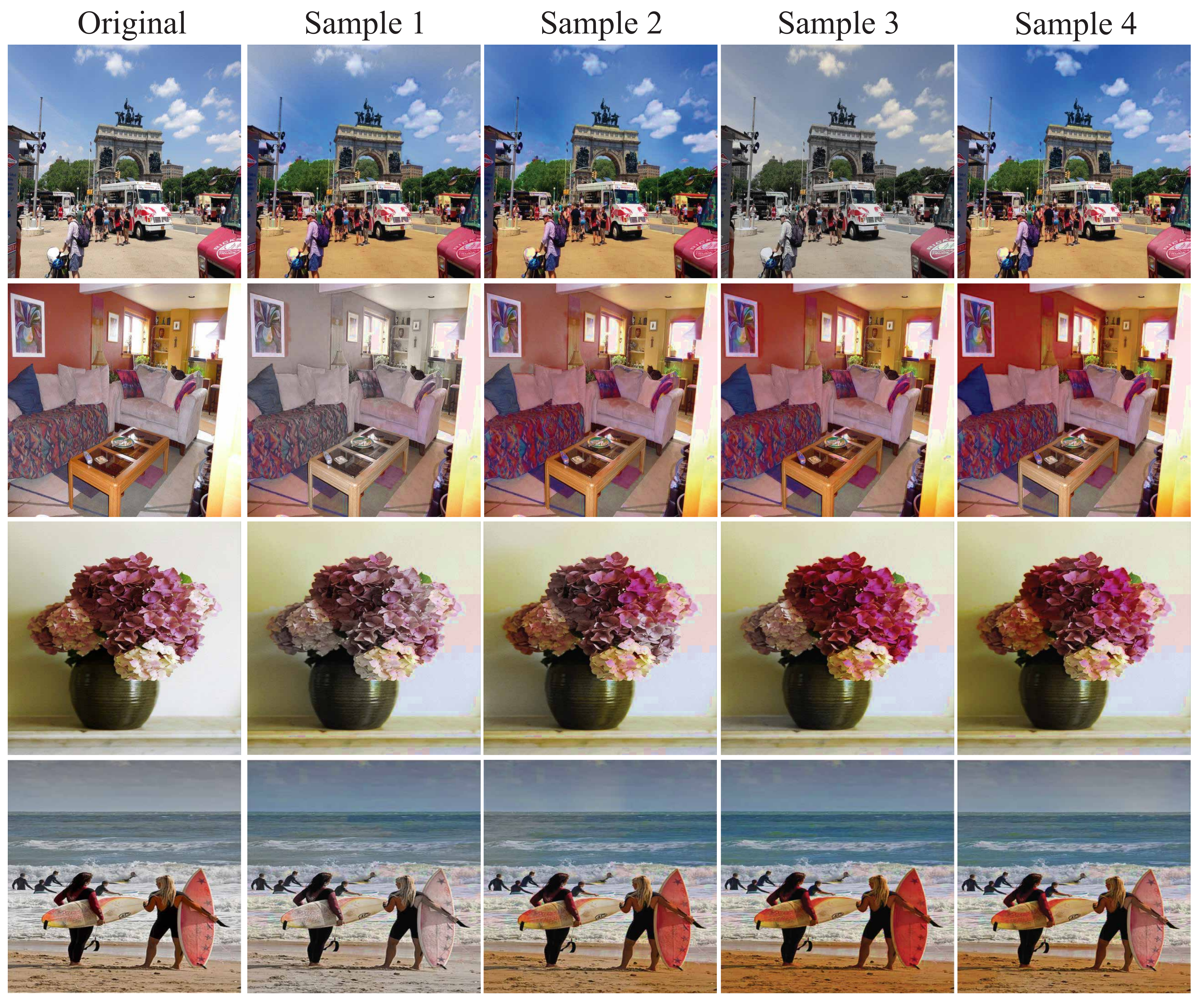}}
  \caption{For each input image in the first column, we sample four different saturation fields from our CDVAE model. 
  Our CDVAE model automatically generates multiple natural saturation adjustments. We learn our conditional distribution from 
   MS-COCO dataset and the outputs show that our prediction respects spatial structure (saturation effects do not bleed 
   across edges) and semantics (objects do not get unnatural or synthetic colors). (Figure best viewed in high resolution)}
  \label{fig:saturation}
\end{figure}

\textbf{Image Resaturation:} More results for image resaturation with our CDVAE (12 gaussian
kernels) and embedding guidance are shown in Figure~\ref{fig:saturation}. For each input image, we draw four 
samples for saturation fields from our conditional model. The diverse saturation adjustment results show 
that our model learns multimodal saturation distributions effectively. Our automatic saturation 
adjustment creates appealing photos. We demonstrate artistic stylization/editing by using our 
automated method.

\begin{figure}[!t]
\centerline{\includegraphics[width=0.5 \textwidth]{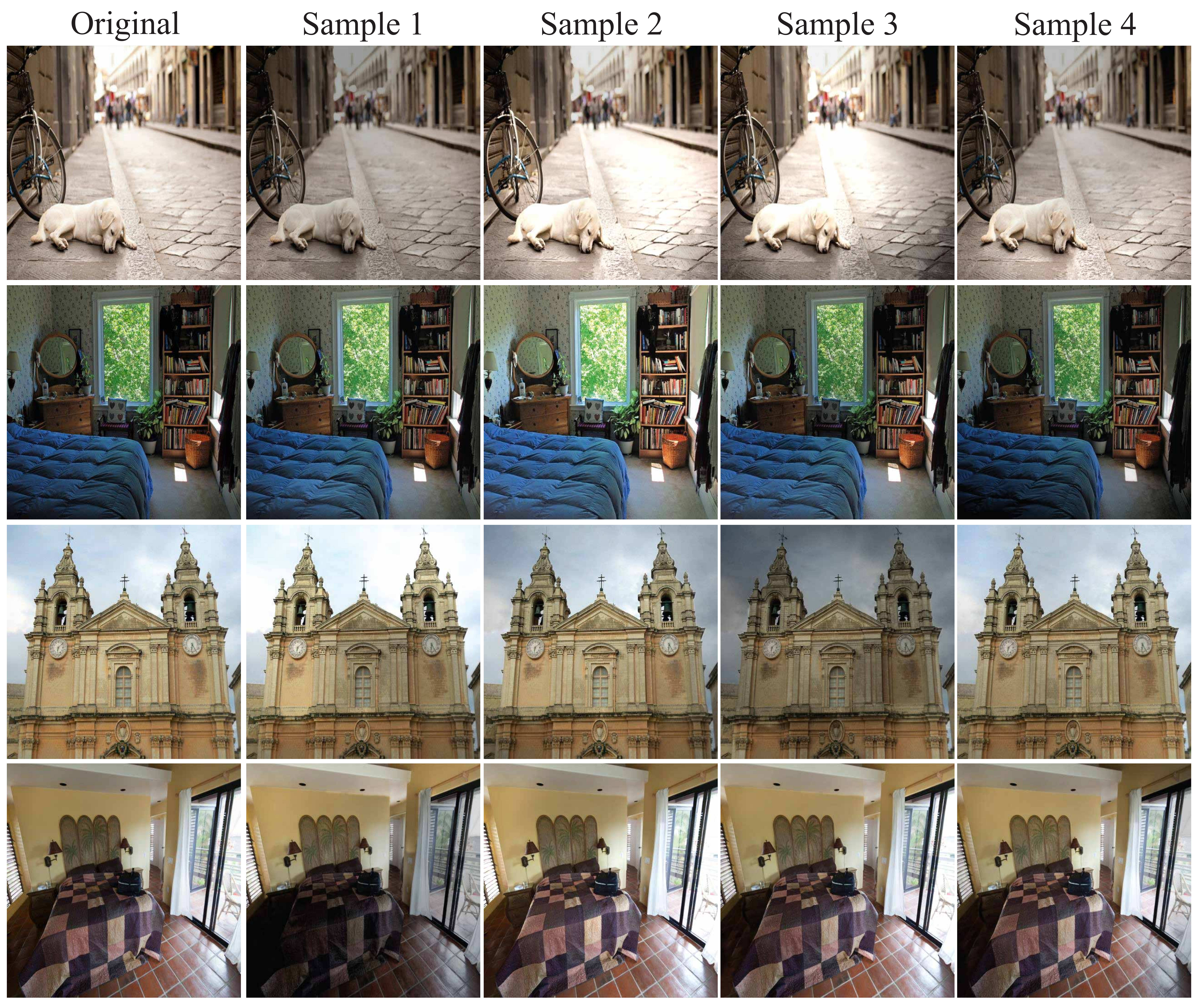}}
  \caption{Input images in the first column are relighted with four samples from our CDVAE model. Since the
   images look natural, CDVAE has automatically learned the potential location of light sources, scene 
   layouts and important objects. This information is critical for correct shading. There is typically no explicit 
   supervision available for these parameters and we show it is not necessary, as CDVAE performs well 
   without needing it. CDVAE learns this implicitly via raw-pixel relationships between albedo and 
   shading. Our sampling creates diverse, yet natural relighted outputs. In the examples, light comes mainly 
   from windows and sky,  and objects are correctly relighted. (Figure best viewed in high resolution)}
  \label{fig:shading}
\end{figure}

\textbf{Photo Relighting:}  In Figure~\ref{fig:shading}, we show additional photo relighting 
results from our method. For each input image, we again draw four samples from the distribution. The 
photo relighting results show that our CDVAE model learns the light source distributions as well 
as important objects. Our model creates light fields coming from reasonable light sources and 
the lighting looks natural, especially on important objects. 

\section{Discussion}

Our CDVAE generates good results qualitatively and quantitatively. However, there are still some 
limitations. Some of them are due to the limitations of VAE based generative models. For example, 
variational auto-encoders and its variants {\bf oversmooth} their outputs, which leads to loss
of spatial structure. Our multilayer gaussian latent variable architecture can capture more 
complex structures, but we do miss out on the finer details compared to ground truth.  
Second, our model -- like all current generative models -- is applied to {\bf low resolution} 
images, meaning that much of the structural and semantic information important to obtaining good 
results is lost. Last, our model has {\bf no spatial hierarchy}. Coarse to fine multiscale 
hierarchy on both generative side and conditional side would likely enable us to produce
results with more details. 

\section{Conclusion}
We described an approach that simplifies building conditional models by building a 
joint model. Our joint model yields a conditional model that can be easily sampled.  This allows us 
to train on scattered data using a joint model.  We have demonstrated our approach on the task
of generating photo-realistic relighted and resaturated images. We propose a metric regularization of code
space which prevents code collapse. In future, this regularization can be investigated in the 
context of other generative models. 

%With all these novel techniques, our CDVAE method brings significant improvements in the diversity and accuracy of the model's output, and creates qualitatively and 
%quantitatively state of the art results. 

\clearpage

\section{Appendix}
\subsection{Architecture Details}
Our CDVAE has a different architecture compared to a CVAE. The detailed architecture of our CDVAE is in 
Table~\ref{architecture}. Write $l_{in}$ for the input layer and $l_{out}$ for the output layer, 
$fc$ stands for a fully connected layer, $mean$ is the mean of the gaussian distribution of the code 
space, and $var$ is the variance of the gaussian distribution of the code space. $Sample$ is 
the process of sampling the gaussian distribution with $mean$ and $var$. $l_{out}$ is sampled from 
$mean_4$ and $var_4$. We use $L_2$ regularization (or weight decay) for the parameters for MDN model. 
The learning rate is set to  $5 \times 10^{-5}$ and we use the ADAM optimizer. We initially set the 
reconstruction cost high, LPP embedding guidance cost high, and MDN cost low. We keep this setting 
and train for 100 epochs. For the next 200 epochs, we gradually decrease the embedding cost, and 
increase the MDN cost. Finally, we keep the relative cost fixed and train another 200 epochs. 

\begin{table*}
\centering
{
\begin{tabular}{ | c | c | c | c | c | c |}
  \hline
   Layers & \multicolumn{2}{|c|}{Conditional DVAE} & $\mbox{MDN}_x$ ($x$ is GMM num) & \multicolumn{2}{|c|}{Generative DVAE} \\
  \hline
  $ l_{in}$ & \multicolumn{2}{|c|}{(None, 1024)} &  & \multicolumn{2}{|c|}{(None, 1024)} \\
  \hline
   $fc1$ & \multicolumn{2}{|c|}{(1024, 512)} &  & \multicolumn{2}{|c|}{(1024, 512)} \\
  \hline
   $activation$ & \multicolumn{2}{|c|}{Leaky Rectify} &  & \multicolumn{2}{|c|}{Leaky Rectify} \\
  \hline
   $fc2$ & \multicolumn{2}{|c|}{(512, 512)} &  & \multicolumn{2}{|c|}{(512, 512)} \\
  \hline
   $activation$ & \multicolumn{2}{|c|}{Leaky Rectify} &  & \multicolumn{2}{|c|}{Leaky Rectify} \\
  \hline
   $fc3_1, fc3_2$ & (512, 64) & (512, 64) &  &  (512, 64) & (512, 64) \\
  \hline
   $activation$ & Identity & SoftPlus &  & Identity & SoftPlus \\
  \hline
   $mean_1, var_1$ & (None, 64) & (None, 64) &  &  (None, 64) & (None, 64) \\
  \hline
   $sample_1$ & \multicolumn{2}{|c|}{(None, 64)}  &  &  \multicolumn{2}{|c|}{(None, 64)} \\
  \hline
   $fc4$ & \multicolumn{2}{|c|}{(64, 256)} &  & \multicolumn{2}{|c|}{(64, 256)} \\
  \hline
   $activation$ & \multicolumn{2}{|c|}{Leaky Rectify} &  & \multicolumn{2}{|c|}{Leaky Rectify} \\
  \hline
   $fc5$ & \multicolumn{2}{|c|}{(256, 256)} &  & \multicolumn{2}{|c|}{(256, 256)} \\
  \hline
   $activation$ & \multicolumn{2}{|c|}{Leaky Rectify} &  & \multicolumn{2}{|c|}{Leaky Rectify} \\
  \hline
   $fc6_1, fc6_2$ & (256, 32) & (256, 32) &  &  (256, 32) & (256, 32) \\
  \hline
   $activation$ & Identity & SoftPlus &  & Identity & SoftPlus \\
  \hline
   $mean_2, var_2$ & (None, 32) & (None, 32) &  &  (None, 32) & (None, 32) \\
  \hline
   $sample_2$ & \multicolumn{2}{|c|}{(None, 32)}  & \begin{tabular}[c]{@{}c@{}}$fc_a(32, (32+1)x)$ = \\ $fc_a(32, (dim(\mu_k)+dim(\pi_k))x)$, \\ activation=tanh, \\ $fc_b((32+1)x,  (32+1)x)$, \\ activation = tanh, \\cost = GMM($z_g|\mu_k, \pi_k, x$)  \end{tabular}  &  \multicolumn{2}{|c|}{(None, 32)} \\
  \hline
   $fc7$ & \multicolumn{2}{|c|}{(32, 256)} &  & \multicolumn{2}{|c|}{(32, 256)} \\
  \hline
   $activation$ & \multicolumn{2}{|c|}{Leaky Rectify} &  & \multicolumn{2}{|c|}{Leaky Rectify} \\
  \hline
   $fc8$ & \multicolumn{2}{|c|}{(256, 256)} &  & \multicolumn{2}{|c|}{(256, 256)} \\
  \hline
   $activation$ & \multicolumn{2}{|c|}{Leaky Rectify} &  & \multicolumn{2}{|c|}{Leaky Rectify} \\
  \hline
   $fc9_1, fc9_2$ & (256, 64) & (256, 64) &  &  (256, 64) & (256, 64) \\
  \hline
   $activation$ & Identity & SoftPlus &  & Identity & SoftPlus \\
  \hline
   $mean_3, var_3$ & (None, 64) & (None, 64) &  &  (None, 64) & (None, 64) \\
  \hline
   $sample_3$ & \multicolumn{2}{|c|}{(None, 64)}  &  &  \multicolumn{2}{|c|}{(None, 64)} \\
  \hline
   $fc10$ & \multicolumn{2}{|c|}{(64, 512)} &  & \multicolumn{2}{|c|}{(64, 512)} \\
  \hline
   $activation$ & \multicolumn{2}{|c|}{Leaky Rectify} &  & \multicolumn{2}{|c|}{Leaky Rectify} \\
  \hline
   $fc11$ & \multicolumn{2}{|c|}{(512, 512)} &  & \multicolumn{2}{|c|}{(512, 512)} \\
  \hline
   $activation$ & \multicolumn{2}{|c|}{Leaky Rectify} &  & \multicolumn{2}{|c|}{Leaky Rectify} \\
  \hline
   $fc12_1, fc12_2$ & (512,1024) & (512,1024) &  & (512,1024) & (512,1024) \\
  \hline
   $activation$ & Identity & SoftPlus &  & Identity & SoftPlus \\
  \hline
   $mean_4, var_4$ & (None, 1024) & (None, 1024) &  &  (None, 1024) & (None, 1024) \\
  \hline
  $ l_{out}$ & \multicolumn{2}{|c|}{(None, 1024)} &  & \multicolumn{2}{|c|}{(None, 1024)} \\
  \hline
\end{tabular}
}
\vspace{2mm}
\label{architecture}
\caption{Details for the CDVAE architecture we proposed. }
\end{table*}

\subsection{DVAE}

The difference between DVAE~\cite{rezende2014stochastic} and VAE~\cite{kingma2013auto} is multiple layers of gaussian latent variables. 
DVAE for $x_c$ (same for $x_g$) consists of $L$ layers of latent variables. To generate a sample from the model, we begin at the top-most 
layer ($L$) by drawing from a Gaussian distribution to get $z_{c,L}$. 
\begin{equation}
P(z_{c,L}) = \mathcal{N}(z_{c,L} | 0, I)
\end{equation}

The mean and variance for the Gaussian distributions at any lower layer is formed by a non-linear transformation of the sample from above layer. 
\begin{equation}
\mu_{c,i} = f_{\mu_{c,i}} (z_{c,i+1})
\end{equation}
\begin{equation}
\sigma_{c,i}^2 = f_{\sigma_{c,i}^2} (z_{c,i+1})
\end{equation}

where $f$ represents multi-layer perceptrons. We descend through the hierarchy by one hot vector sample process.
\begin{equation}
z_{c,i} = \mu_{c,i} + \xi_i \sigma_{c,i}
\end{equation}

where $\xi_i$ are mutually independent Gaussian variables. $x_c$ is generated by sampling from the Gaussian distribution at the lowest layer.
\begin{equation}
P(x_c|z_{c,1}) = \mathcal{N}(x_c|\mu_{c,0}, \sigma_{c,0}^2)
\end{equation}

The joint probability distribution $P(x_c, z_c)$ of this model is formulated as
\begin{equation}
\begin{split}
P (x_c, z_c) & = P(x_c|z_{c,1}) P(z_c) \\
& = P(x_c|z_{c,1}) P (z_{c,L}) \prod_{i=1}^{L-1} P (z_{c,i} | z_{c,i+1})
\end{split}
\end{equation}
where $P(z_{c,i} | z_{c,i+1}) = \mathcal{N} (z_{c,i} | \mu_{c,i}, \sigma_{c,i}^2) $. Other details of the DVAE model are similar to VAE. 

\subsubsection{Inference}
DVAE with several layers of dependent stochastic variables are difficult to train which limits the improvements obtained using these
highly expressive models. LVAE~\cite{sonderby2016ladder} recursively corrects the generative distribution by a data dependent approximate
likelihood in a process resembling the recent Ladder Network. It utilizes a deeper more distributed hierarchy of latent variables and captures more
complex structures. We follow this work and for $x_c$, write $\mu_{c,p,i}$ and $\sigma_{c,p,i}^2$ for the mean and variance on the $i$'s
level of generative side, write $\mu_{c,q,i}$ and $\sigma_{c,q,i}^2$ for the mean and variance on the $i$'s level of inference side. 

This changes the notation in the previous part on the generative side. 
\begin{equation}
P_{p}(z_c) = P_{p}(z_{c,L}) \prod_{i=1}^{L-1} P_{p}(z_{c,i}|z_{c,i+1})
\end{equation}

\begin{equation}
P_{p}(z_{c,L}) = \mathcal{N}(z_{c,L} | 0, I)
\end{equation}

\begin{equation}
P_{p}(z_{c,i}|z_{c,i+1}) = \mathcal{N}(z_{c,i}|\mu_{c,p,i}, \sigma_{c,p,i}^2)
\end{equation}

\begin{equation}
P_{p}(x_c | z_{c,1}) = \mathcal{N}(x_c | \mu_{c,p,0}, \sigma_{c,p,0}^2)
\end{equation}

On the inference side, the notation also changes.
\begin{equation}
P_{q}(z_c|x_c) = P_{q}(z_{c,1}|x_c) \prod_{i=2}^{L} P_{q}(z_{c,i}|z_{c,i-1})
\end{equation}

\begin{equation}
P_{q}(z_{c,1}|x_c) = \mathcal{N}(z_{c,1} | \mu_{c,q,1}, \sigma_{c,q,1}^2)
\end{equation}

\begin{equation}
P_{q}(z_{c,i}|z_{c,i-1}) = \mathcal{N}(z_{c,i}|\mu_{c,q,i}, \sigma_{c,q,i}^2)
\end{equation}

During inference, first a deterministic upward pass computes the approximate distribution $\hat{\mu}_{c,q,i}$ and $\hat{\sigma}_{c,q,i}^2$. 
This is followed by a stochastic downward pass recursively computing both the approximate posterior and generative distributions.
\begin{equation}
P_{q}(z_c|x_c) = P_{q}(z_{c,L}|x_c) \prod_{i=1}^{L-1} P_{q}(z_{c,i}|z_{c,i+1})
\end{equation}

\begin{equation}
\sigma_{c,q,i} = \frac{1}{\hat{\sigma}_{c,q,i}^{-2} + \sigma_{c,p,i}^{-2}}
\end{equation}

\begin{equation}
\mu_{c,q,i} = \frac{\hat{\mu}_{c,q,i} \hat{\sigma}_{c,q,i}^{-2} + {\mu}_{c,p,i} {\sigma}_{c,p,i}^{-2}}{\hat{\sigma}_{c,q,i}^{-2} + {\sigma}_{c,p,i}^{-2}}
\end{equation}

\begin{equation}
P_{q}(z_{c,i}|\cdot) = \mathcal{N} (z_{c,i}|\mu_{c,q,i}, \sigma_{c,q,i}^2)
\end{equation}
where $\mu_{c,q,L} = \hat{\mu}_{c,q,L}$ and $\sigma_{c,q,L}^2 = \hat{\sigma}_{c,q,L}^2$.

\subsection{Joint Models}
First, we prove that if the joint probability is independent, we will get two separate DVAEs. Then, we prove the derivations for joint model with non-independent joint probability. 

\subsubsection{Separate DVAEs}
From Section 3 in the paper, the joint probability $P(x_c, x_g)$ in CDVAE model is
\begin{equation}
P(x_c, x_g) = \int_z P(x_c|z_c) P(x_g|z_g) P(z_g, z_c) dz_g dz_c
\label{eq:joint}
\end{equation}

If $z_c$ and $z_g$ are independent, so $P(z_g, z_c) = P(z_g) P(z_c)$, and Equation~\ref{eq:joint} can be transformed
\begin{equation}
\begin{split}
P(x_c, x_g) & = \int_z P(x_c|z_c) P(x_g|z_g) P(z_g) P(z_c) dz_g dz_c \\
& = \int_z (P(x_c|z_c) P(z_c)) dz_c (P(x_g|z_g) P(z_g)) dz_g \\
& = \int_{z_c} P(x_c|z_c) P(z_c) dz_c + \int_{z_g} P(x_g|z_g) P(z_g) dz_g \\
& = P(x_c) + P(x_g)
\end{split}
\end{equation}

where $P(x_c)$ is DVAE model for $x_c$ and $P(x_g)$ is DVAE model for $x_g$.

\subsubsection{Joint Model Derivation}
From Section 2 in the paper, we have objective function for VAE as
\begin{equation}
\mbox{VAE}(\theta) = \sum_{data}[\mathbb{E}_Q \log P(x|z) - \mathbb{D}(Q || P(z))]
\end{equation}

where $Q=P(z|x)$. Applying the same derivations, the objective function for our CDVAE model can be written as
\begin{equation}
\resizebox{1.0\hsize}{!}{$ \mbox{CDVAE}(\theta_c, \theta_g) = \sum_{data}[\mathbb{E}_Q \log P(x_c, x_g|z_c, z_g) - \mathbb{D}(Q || P(z_c, z_g))] $}
\label{eq:cdvae}
\end{equation}

where $Q=P(z_c, z_g|x_c, x_g)$. Assume it is possible to encode $x_c$ without seeing $x_g$, then the variational distribution 
$Q=P(z_c|x_c) P(z_g|x_g)$ applies. It is also possible to decode $x_c$ without seeing $x_g$, so we have 
$P(x_c, x_g | z_c, z_g) = P(x_c | z_c) P(x_g | z_g)$. With these formulas, Equation~\ref{eq:cdvae} can be transformed
\begin{equation}
\begin{split}
\mathbb{E}_Q \log P(x_c, x_g|z_c, z_g) & = \mathbb{E}_Q \log (P(x_c|z_c)P(x_g|z_g)) \\
& = \mathbb{E}_{Q_1} \log P(x_c|z_c) \\
& + \mathbb{E}_{Q_2} \log P(x_g|z_g)
\end{split}
\end{equation}

where $Q_1=P(z_c|x_c)$ and $Q_2=P(z_g|x_g)$. The joint distribution can be written as $\log P(z_c, z_g) = \log P(z_c) + \log P(z_g) + F_{\mbox{mdn}}(z_c, z_g)$, 
so we have the following equations for the second part. 
\begin{equation}
\begin{split}
\mathbb{D}(Q || P(z_c, z_g)) & = \mathbb{D}(P(z_c|x_c) P(z_g|x_g) || P(z_c, z_g)) \\
& = \mathbb{E}_Q(\log (P(z_c|x_c) P(z_g|x_g)) - \log P(z_c, z_g)) \\
& = \mathbb{E}_{Q_1}(\log P(z_c|x_c)) + \mathbb{E}_{Q_2}(\log P(z_g|x_g)) \\
& - \mathbb{E}_{Q_1}(\log P(z_c)) - \mathbb{E}_{Q_2}(\log P(z_g)) \\
& - \mathbb{E}_Q(F_{\mbox{mdn}}(z_c, z_g)) \\
& = \mathbb{D}(Q_1 || P(z_c)) + \mathbb{D}(Q_2 || P(z_g))\\
& - \mathbb{E}_Q(F_{\mbox{mdn}}(z_c, z_g))
\end{split}
\end{equation}

In our CDVAE model, we have $\log P(z_c) = -\frac{z_c^T z_c}{2}$ and $\log P(z_g) = -\frac{z_g^T z_g}{2}$ because $z_c$ and $z_g$ are Gaussian distributions. Our
CDVAE objective function turns into
\begin{equation}
\resizebox{1.0\hsize}{!}{$ \mbox{CDVAE}(\theta_c, \theta_g) = \mbox{DVAE}(\theta_c) + \mbox{DVAE}(\theta_g) + \sum_{data} \mathbb{E}_Q(F_{\mbox{mdn}}(z_c, z_g)) $}
\end{equation}

\subsection{Embedding Influence}
\begin{figure*}
\centerline{  \includegraphics[width=0.75 \textwidth]{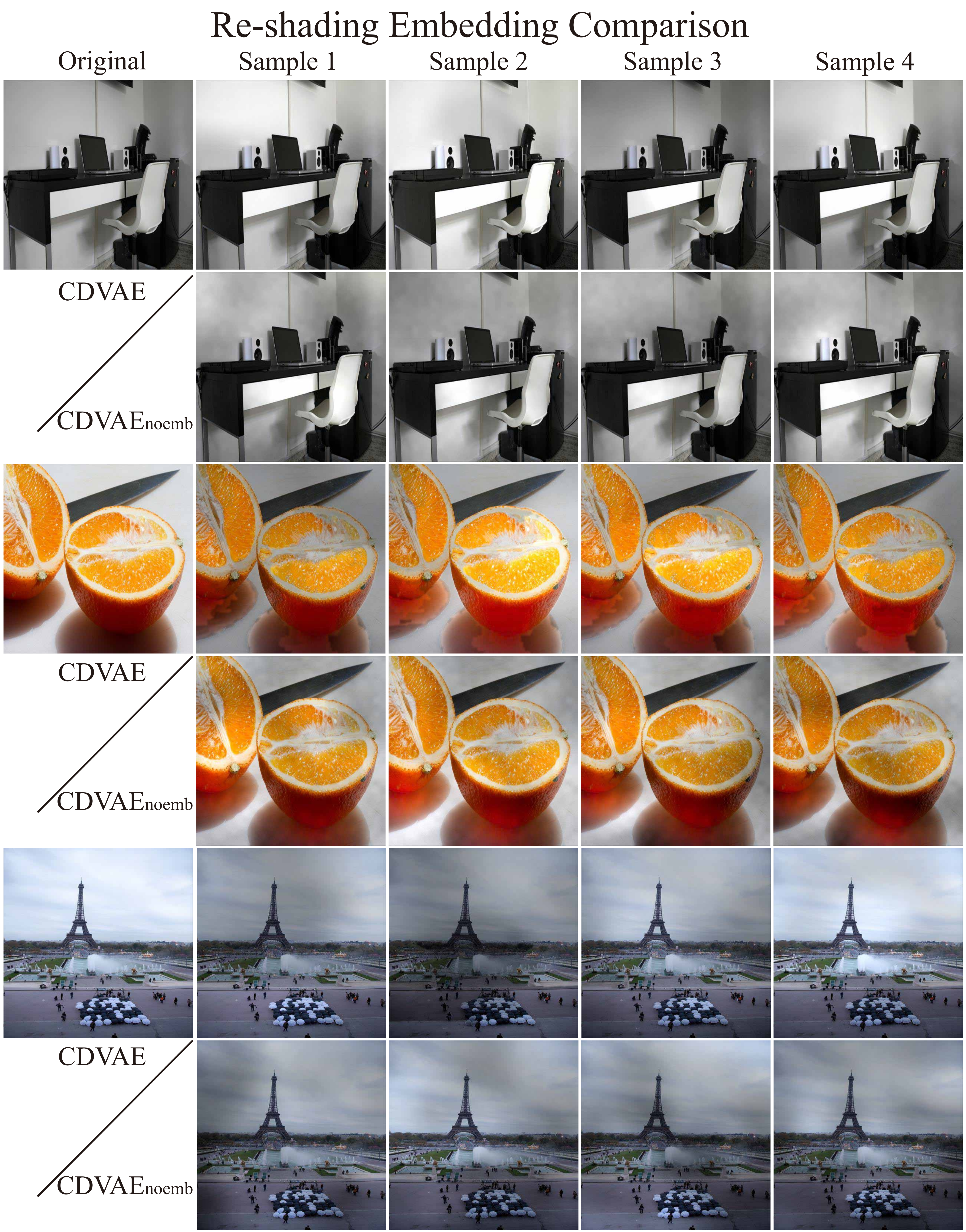}}
  \caption{Comparisons to no embedding guidance for re-shading results (part 2). The re-shading results without embedding guidance tend to have less variety, more flaws and artifacts. }
  \label{fig:cmpno2}
\end{figure*}

\begin{figure*}
\centerline{  \includegraphics[width=0.75 \textwidth]{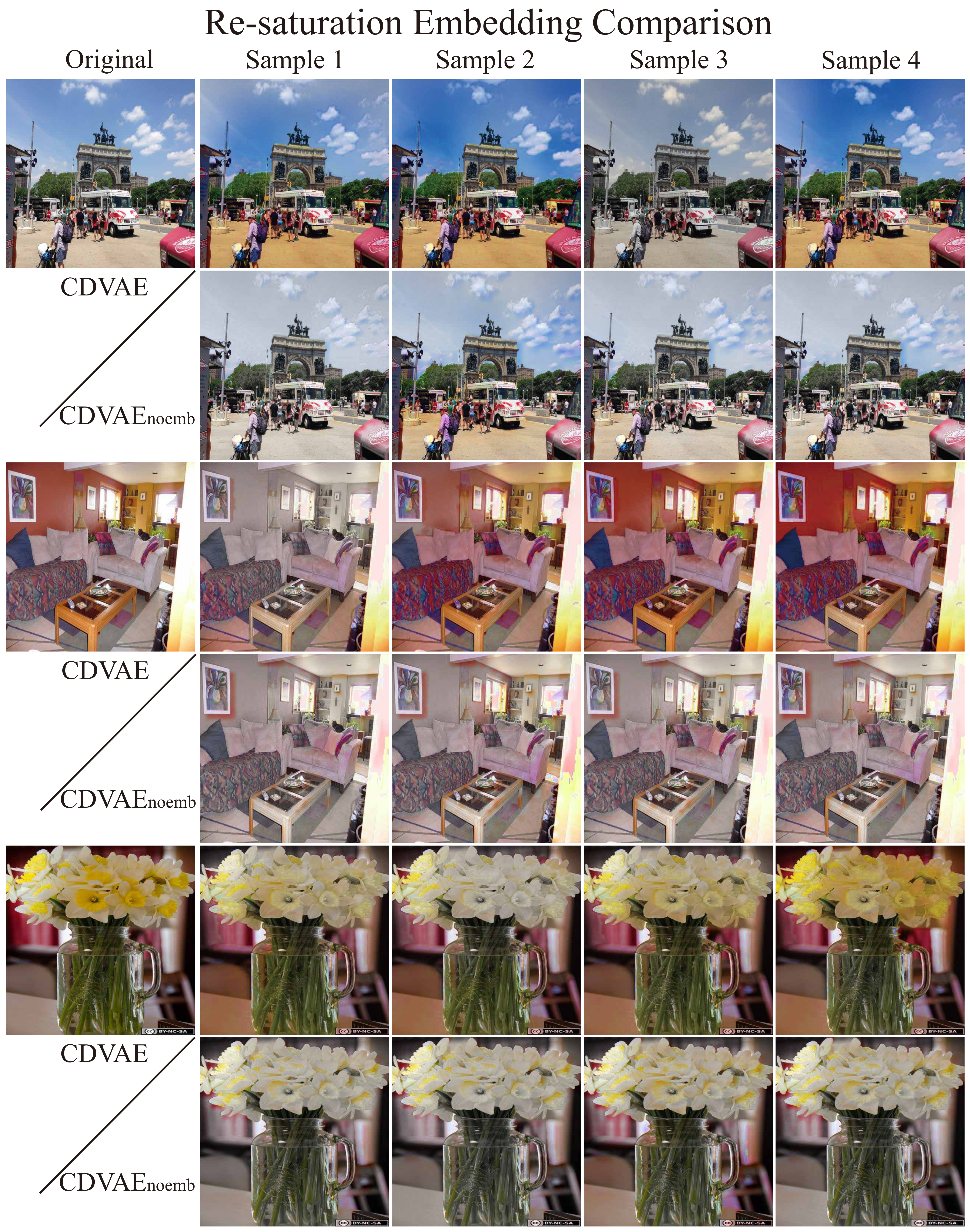}}
  \caption{Comparisons to no embedding guidance for re-saturation results (part 1). The re-saturation results without
embedding guidance tend to have limited variety and produce less vivid results. }
  \label{fig:cmpnosat1}
\end{figure*}

We compare the results with embedding guidance and without embedding guidance. The comparisons for re-shading can be found in Figures~\ref{fig:cmpno2}. The re-shading results without embedding guidance tend to have less variety, more flaws and artifacts. 
The comparisons for re-saturation can be found in Figures~\ref{fig:cmpnosat1}. The re-saturation results without
embedding guidance tend to have limited variety and produce less vivid results. 

\subsection{Quantitative Results}
The detailed quantitative evaluation results for \textbf{photo relighting} are in
Table~\ref{tb:shading} and  \textbf{image resaturation} are in Table~\ref{tb:saturation}.
The tables contain best error to ground-truth with different sample numbers. As the sample 
number increases, the error drops fast at beginning, and then becomes stable. Our CDVAEs 
are consistently better than other methods. The second parts of both tables are average 
variances across 100 samples. We only report the final variance, since it almost does not change 
with the sample number. The variance we report comes from 100 samples. 

\begin{table*}
\centering
{
\begin{tabular}{ | c | c | c | c | c | c | c |}
  \hline
   & \multicolumn{5}{|c|}{Best Error to Ground Truth} & Variance \\
  \hline
    & Sample\# 3 & Sample\# 10 & Sample\# 30 & Sample\# 60 & Sample\# 100 & Sample\# \\
  \hline
  NN   & 3.04 & 2.30 & 1.93 & 1.76 & 1.66 & 1.61 \\
  \hline
  CVAE  & 2.07 & 1.83 & 1.68 &  1.60 &  1.56 & 0.19 \\
  \hline
  CGAN & 3.07 & 2.49 & 2.16 & 2.02 & 1.94 & 1.19 \\
  \hline
  CPixel & 3.06 & 2.32 & 1.91 & 1.74 & 1.59 & 1.92 \\
  \hline
  $\mbox{CDVAE}_{noemb}$ & 2.78 & 2.19 & 1.82 & 1.66 & 1.57 & 1.39\\
  \hline
  CDVAE4 & 2.44 & 1.66 & 1.33 & 1.20 & 1.11 & 1.77 \\
  \hline
  CDVAE12 & 2.49 & 1.69 & 1.33 & 1.20 & 1.12 & 1.74 \\
  \hline
\end{tabular}
}
\vspace{2mm}
\caption{Photo relighting results. First part is best error to ground truth with different sample numbers; second part is variance, which 
is stable with different sample numbers. (all results need $\times 10^{-2}$)}
\label{tb:shading}
\end{table*}

\begin{table*}
\centering
{
\begin{tabular}{ | c | c | c | c | c | c | c |}
  \hline
   & \multicolumn{5}{|c|}{Best Error to Ground Truth} & Variance \\
  \hline
    & Sample\# 3 & Sample\# 10 & Sample\# 30 & Sample\# 60 & Sample\# 100 & Sample\# \\
  \hline
  NN  & 10.12 & 8.40 & 7.09 & 6.52 & 6.20 & 4.58 \\
  \hline
  CVAE  & 6.73 & 5.59 & 4.93 &  4.53 &  4.25 & 1.20 \\
  \hline
  CGAN & 8.06 & 6.20 & 5.37 & 5.03 & 4.83 & 4.79\\
  \hline
  CPixel & 7.94 & 6.43 & 5.94 & 5.51 & 5.29 & 4.34\\
  \hline
  $\mbox{CDVAE}_{noemb}$ & 7.08 & 5.74 & 4.95 & 4.57 & 4.32 & 1.53\\
  \hline
  MDN4 & 6.62 & 5.11 & 4.37 & 4.05 & 3.86 & 3.48 \\
  \hline
  MDN12 & 6.40 & 5.04 & 4.33 & 4.02 & 3.82 & 3.55 \\
  \hline
\end{tabular}
}
\vspace{2mm}
\caption{Image re-saturation results. First part is best error to ground truth with different sample numbers; second part is variance, which 
is stable with different sample numbers. (all results need $\times 10^{-2}$)}
\label{tb:saturation}
\end{table*}

\subsection{Qualitative Results}
\begin{figure*}
\centerline{  \includegraphics[width=0.75 \textwidth]{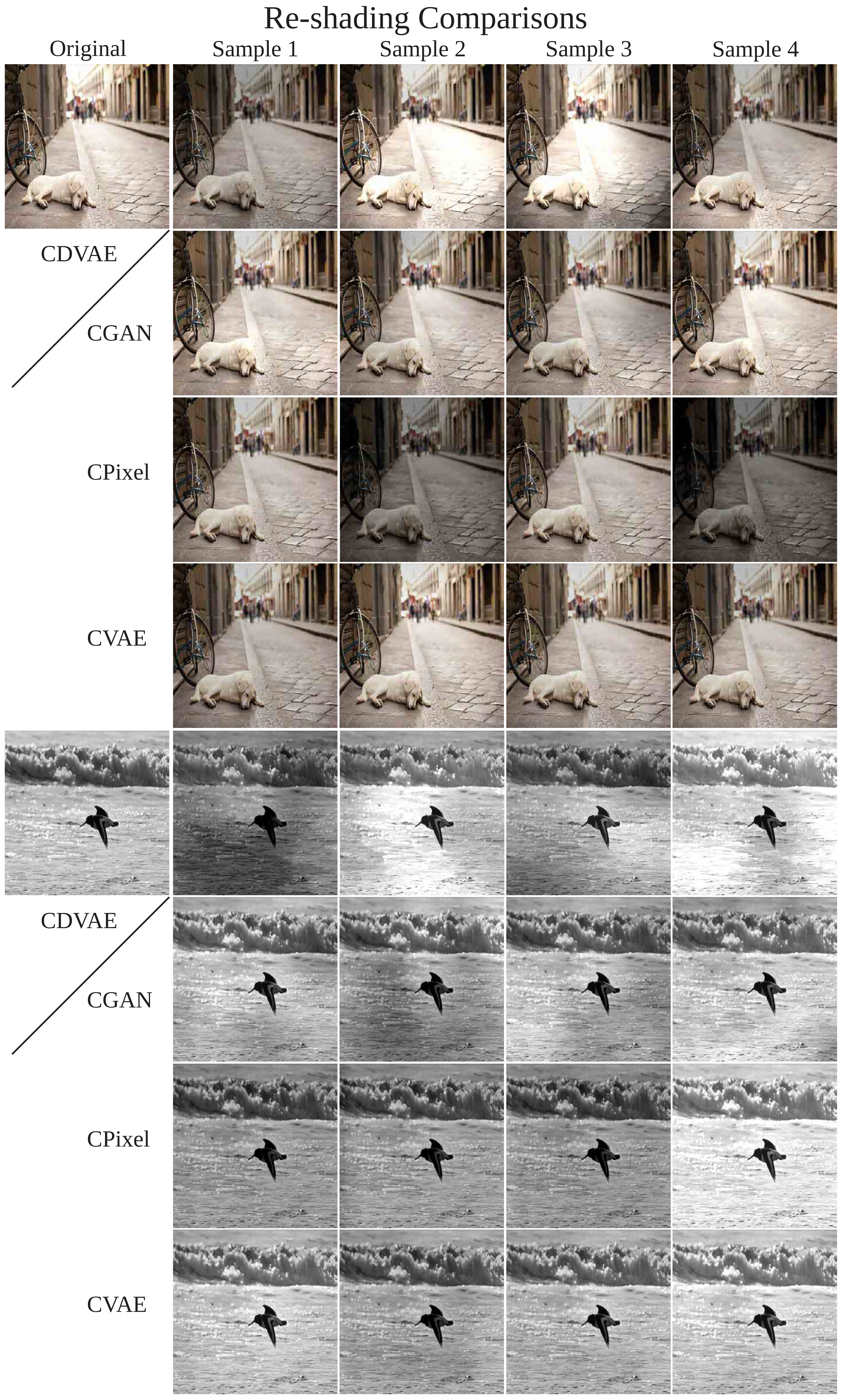}}
  \caption{Photo relighting results (part 1). Photo relighting results with CGAN tend to have less variety and be less reasonable;
results with CPixel tend to be extreme and random, and they also have less spatial structures; results with CVAE suffers from mode collapsion and have limited variety.}
  \label{fig:r01}
\end{figure*}

\begin{figure*}
\centerline{  \includegraphics[width=0.75 \textwidth]{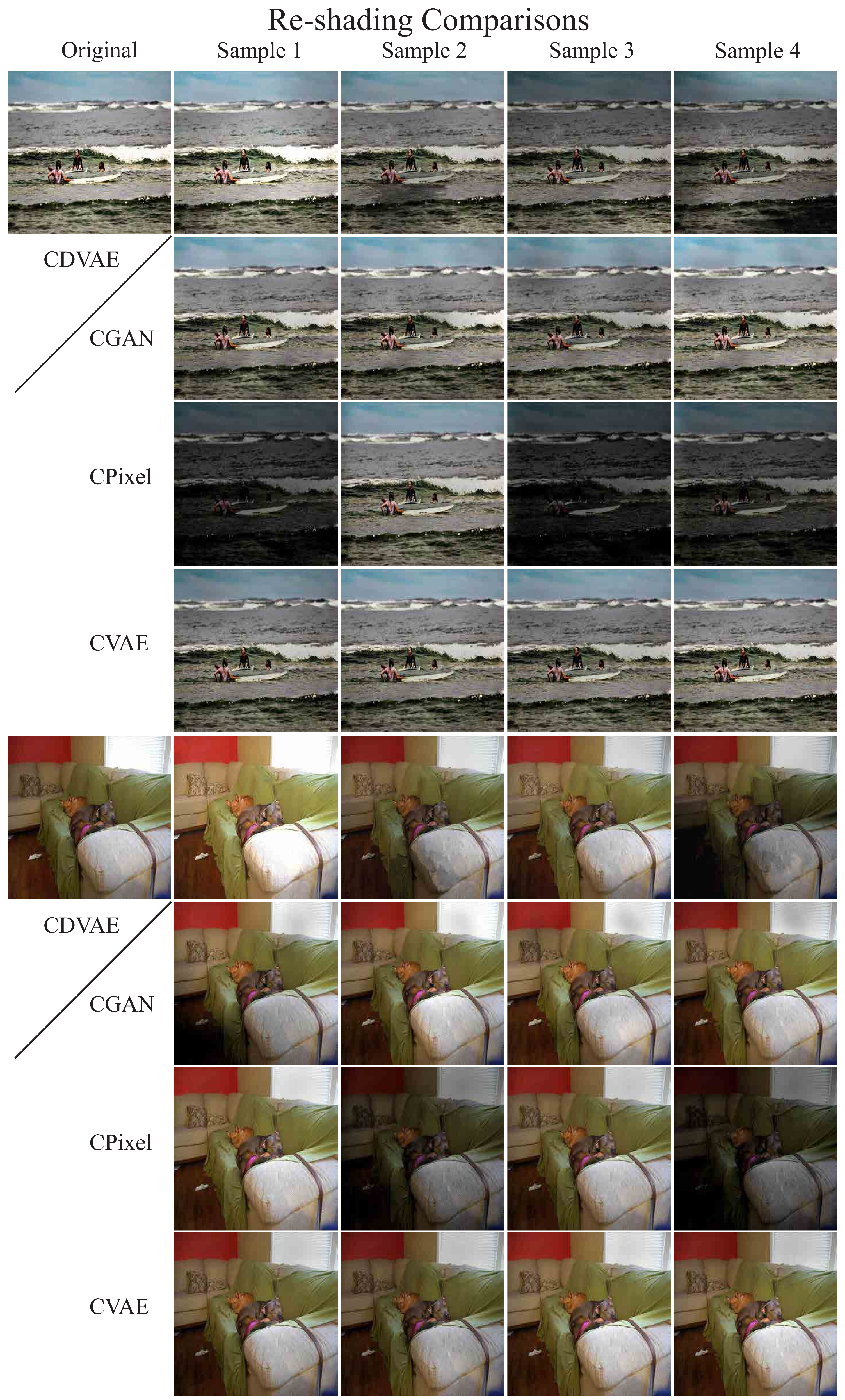}}
  \caption{Photo relighting results (part 3). Photo relighting results with CGAN tend to have less variety and be less reasonable;
results with CPixel tend to be extreme and random, and they also have less spatial structures; results with CVAE suffers from mode collapsion and have limited variety.}
  \label{fig:r11}
\end{figure*}

\begin{figure*}
\centerline{  \includegraphics[width=0.75 \textwidth]{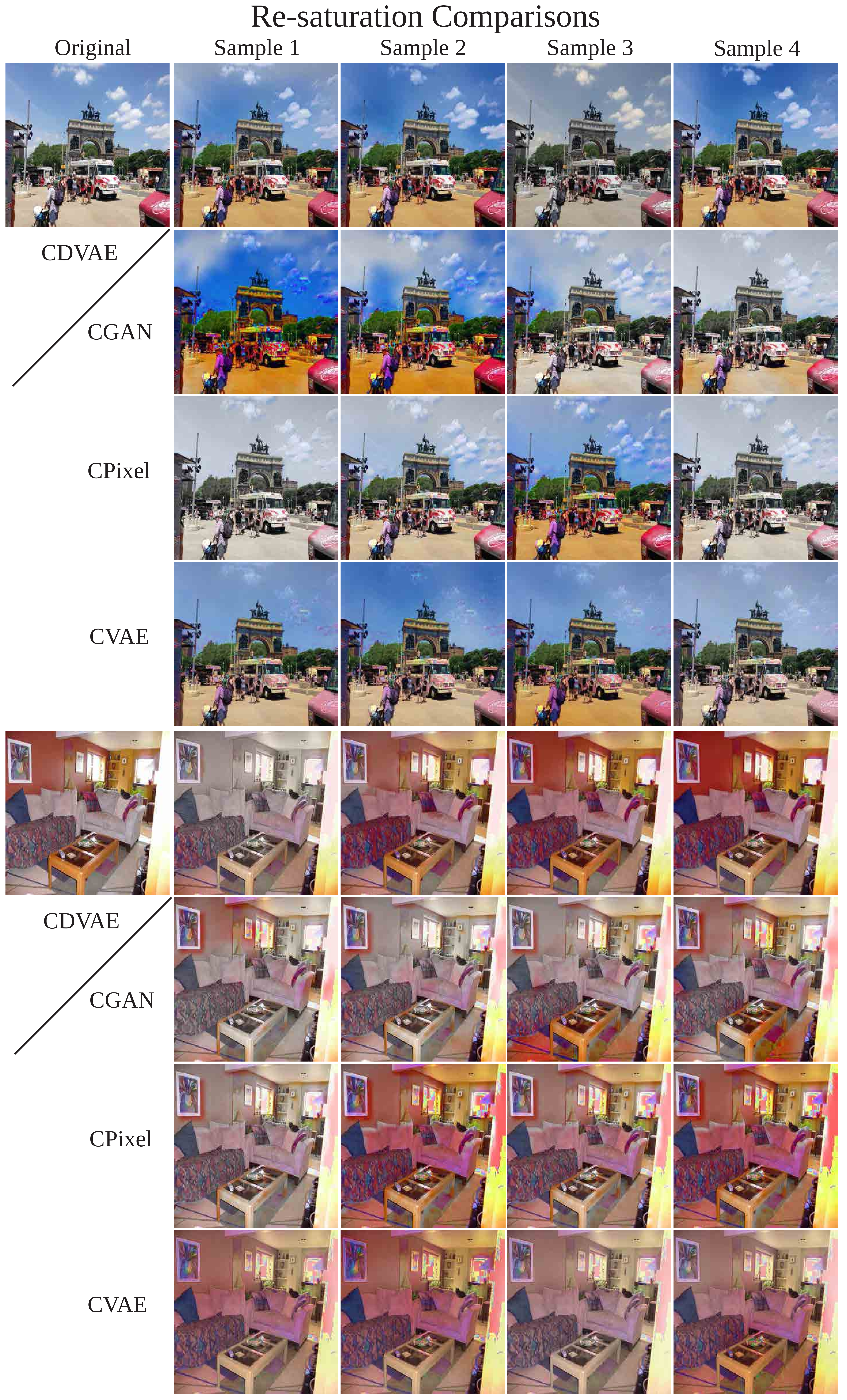}}
  \caption{Image re-saturation results (part 1). Image re-saturation results with CGAN tend to ignore the image content and like random,
and creates various of artifacts; results with CPixel tend to be extreme, and either like random or go into mode collapsion; results with CVAE have limited variety and creates
more artifacts. }
  \label{fig:s01}
\end{figure*}

\begin{figure*}
\centerline{  \includegraphics[width=0.75 \textwidth]{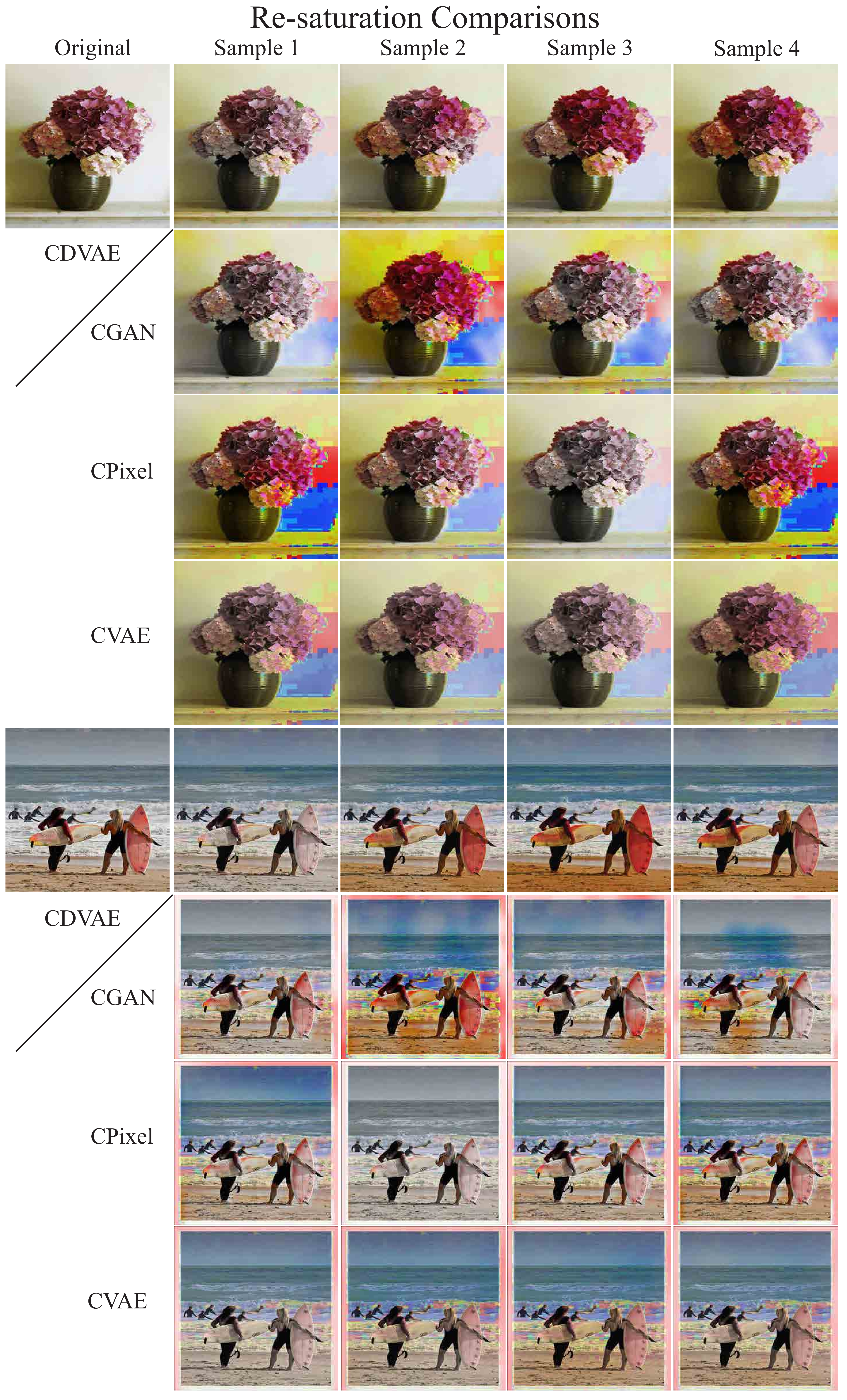}}
  \caption{Image re-saturation results (part 2). Image re-saturation results with CGAN tend to ignore the image content and like random,
and creates various of artifacts; results with CPixel tend to be extreme, and either like random or go into mode collapsion; results with CVAE have limited variety and creates
more artifacts. }
  \label{fig:s02}
\end{figure*}

We include more qualitative results and comparisons in this section. \textbf{Photo relighting} results and comparisons can be found in Figure~\ref{fig:r01},
~\ref{fig:r11}. Photo relighting results with CGAN tend to have less variety and be less reasonable;
results with CPixel tend to be extreme and random, and they also have less spatial structures; results with CVAE suffers from mode collapsion and have limited variety.
\textbf{Image re-saturation} results and comparisons can be found in Figure~\ref{fig:s01},~\ref{fig:s02}. 
Image re-saturation results with CGAN tend to ignore the image content and like random,
and creates various of artifacts; results with CPixel tend to be extreme, and either like random or go into mode collapsion; results with CVAE have limited variety and creates
more artifacts. 

\clearpage 

\clearpage

\clearpage

\clearpage
\clearpage

\clearpage
\clearpage

{\small
\bibliographystyle{ieee}
\bibliography{egbib}
}

\end{document}